\documentclass[letterpaper]{article} 
\usepackage[table]{xcolor}
\usepackage{aaai2026}  
\usepackage{times}  
\usepackage{helvet}  
\usepackage{courier}  
\usepackage[hyphens]{url}  
\usepackage{graphicx} 
\usepackage{natbib}  
\usepackage{caption} 
\usepackage{multirow}
\usepackage{adjustbox}
\usepackage{booktabs}
\usepackage{tabularx}
\usepackage{makecell}
\usepackage{pifont}
\usepackage{textcomp}
\usepackage{algorithm}
\usepackage{algorithmic}

\usepackage{amsmath}
\usepackage{threeparttable}
\usepackage{bbding}
\usepackage{amssymb}
\usepackage{url}
\usepackage{pdfpages}

\usepackage{siunitx}
\definecolor{headercolor}{gray}{0.93}
\definecolor{teal}{RGB}{0, 128, 128}
\definecolor{lightcoral}{RGB}{240, 128, 128}

\newcommand{\cmark}{\ding{51}} 
\newcommand{\xmark}{\ding{55}} 
\newcommand{\yes}{\textcolor{teal}{\cmark}}
\newcommand{\no}{\textcolor{lightcoral}{\xmark}}

\definecolor{bestcolor}{HTML}{E8F5E9}   
\definecolor{secondcolor}{HTML}{FFF9C4} 
\definecolor{rowgray}{HTML}{F5F5F5}     

\usepackage{newfloat}
\usepackage{listings}
\DeclareCaptionStyle{ruled}{labelfont=normalfont,labelsep=colon,strut=off} 
\floatstyle{ruled}
\newfloat{listing}{tb}{lst}{}
\floatname{listing}{Listing}
\usepackage{xspace}
\newcommand{\modelname}{\textsc{ForestLLM}\xspace}

\title{{\modelname}: Large Language Models Make Random Forest Great \\ on Few-shot Tabular Learning}
\author {
    Zhihan Yang\textsuperscript{\rm 1}\thanks{Equal contribution.}\thanks{Work done during an internship at Microsoft Research Asia.},
    Jiaqi Wei\textsuperscript{\rm 2}\footnotemark[1],
    Xiang Zhang\textsuperscript{\rm 3},
    Haoyu Dong\textsuperscript{\rm 4},
    Yiwen Wang\textsuperscript{\rm 5},
    Xiaoke Guo\textsuperscript{\rm 2}, \\
    Pengkun Zhang\textsuperscript{\rm 6},
    Yiwei Xu\textsuperscript{\rm 7},
    Chenyu You\textsuperscript{\rm 8}
}
\affiliations {
    \textsuperscript{\rm 1} National University of Singapore 
    \textsuperscript{\rm 2} Zhejiang University
    \textsuperscript{\rm 3} University of British Columbia 
    \textsuperscript{\rm 4} Microsoft \\
    \textsuperscript{\rm 5} University of Science and Technology of China
    \textsuperscript{\rm 6} South China University of Technology \\
    \textsuperscript{\rm 7} Cisco
    \textsuperscript{\rm 8} Stony Brook University
}

\usepackage{bibentry}

\begin{document}

\maketitle

\begin{abstract}
Tabular data high-stakes critical decision-making in domains such as finance, healthcare, and scientific discovery. Yet, learning effectively from tabular data in few-shot settings, where labeled examples are scarce, remains a fundamental challenge. Traditional tree-based methods often falter in these regimes due to their reliance on statistical purity metrics, which become unstable and prone to overfitting with limited supervision. At the same time, direct applications of large language models (LLMs) often overlook its inherent structure, leading to suboptimal performance.
To overcome these limitations, we propose {\modelname}, a novel framework that unifies the structural inductive biases of decision forests with the semantic reasoning capabilities of LLMs. Crucially, {\modelname} leverages the LLM only during training, treating it as an {\emph{offline model designer}} that encodes rich, contextual knowledge into a lightweight, interpretable forest model, eliminating the need for LLM inference at test time.
Our method is two-fold.
First, we introduce a \textit{semantic splitting criterion} in which the LLM evaluates candidate partitions based on their coherence over \textbf{both labeled and unlabeled data}, enabling the induction of more robust and generalizable tree structures under few-shot supervision. 
Second, we propose a one-time in-context inference mechanism for \textit{leaf node stabilization}, where the LLM distills the decision path and its supporting examples into a concise, deterministic prediction, replacing noisy empirical estimates with semantically informed outputs.
Across a diverse suite of few-shot classification and regression benchmarks, {\modelname} achieves state-of-the-art performance. These results highlight the promise of using LLMs as {offline model designers}, rather than online predictors, for scalable and efficient tabular learning under data scarcity. Our code is available at \url{https://github.com/kelvin715/ForestLLM}.


\end{abstract}

\section{Introduction}


Tabular data, structured as collections of heterogeneous, feature-label pairs, is central to decision-making in critical domains such as healthcare, finance, and public policy~\citep{hernandez2022synthetic,frosch2010decade,assefa2020generating,Johnson2016MIMICIIIAF,ulmer2020trust,arun2016loan,chen2023machine,zhang2025tokenization}. However, building reliable models from such data in the few-shot regime remains a longstanding challenge. Unlike images or text, tabular data lacks spatial or sequential structure, and its highly combinatorial, non-Euclidean nature undermines the geometric assumptions underlying most modern few-shot learning techniques~\citep{chen2018closer,majumder2022few,nam2023stunt}.
As a result, these methods often perform poorly on tabular tasks with limited supervision.

Tree-based models such as Random Forests and Gradient Boosted Trees~\citep{chen2016xgboost,ke2017lightgbm,prokhorenkova2018catboost} remain strong performers on tabular benchmarks, thanks to their inductive bias toward hierarchical data partitioning. Yet under low-data conditions, their reliability degrades. Conventional split selection heuristics, like information gain or Gini impurity, rely entirely on labeled data and become unstable when estimates are noisy. Moreover, predictions at the leaves are often based on only a handful of examples, resulting in high variance and poor calibration. These issues are further exacerbated by the fact that unlabeled data, though often available, is ignored entirely during training.


\begin{table}[!tbp]
  \centering
  \caption{Comparison of LLM-based Tabular Methods.}
  \renewcommand{\arraystretch}{1.4}
  \resizebox{\linewidth}{!}{%
  \begin{tabular}{l|c|c|cc}
    \toprule
    \multirow{2}{*}{\textbf{Method}} & 
    \multirow{2}{*}{\textbf{\makecell[c]{No LLM\\Fine-tuning Needed}}} & 
    \multirow{2}{*}{\textbf{\makecell[c]{No LLM\\Needed at Inference}}} & 
    \multicolumn{2}{c}{\textbf{Task Support}} \\
    & & & \textbf{Classification} & \textbf{Regression} \\
    \midrule
    LIFT (NeurIPS\textquotesingle22)  & \no  & \no  & \yes & \yes  \\
    TabLLM (AISTATS\textquotesingle23) & \no  & \no  & \yes & \no  \\
    FeatLLM (ICML\textquotesingle24) & \yes & \yes & \yes & \no \\
    TP-BERTa (ICLR\textquotesingle24) & \yes & \no  & \yes & \yes \\
    P2T (COLM\textquotesingle24) & \yes & \no  & \yes & \yes \\
    \rowcolor{teal!15} \textbf{{\modelname} (Ours)} & \yes & \yes & \yes & \yes \\
    \bottomrule
  \end{tabular}
  }
  \label{tab:llm_comparison}
  \vspace{-1em}
\end{table}

In parallel, the generalization and reasoning capabilities of Large Language Models (LLMs) have spurred interest in applying them to tabular learning~\citep{dinh2022lift,hegselmann2023tabllm,han2024large,yan2024making,nam2024tabular,wen2024supervised,zhang2025prompt,yan2025small}. But most existing approaches serialize rows into text and rely on prompting or fine-tuning, discarding the structural regularities that classical models are designed to exploit. Despite their flexibility, such methods frequently underperform tree ensembles on core supervised tasks, particularly in few-shot settings~\citep{grinsztajn2022why,borisov2022deep,mcelfresh2023neural,zabergja2024deep}.

In this paper, we explore a new direction: \textit{rather than replacing classical tabular models with foundation models, can we instead enlist LLMs as offline architects, guiding model design during training but removed entirely at test time? Can we use LLMs to imbue decision trees with semantic inductive priors, while preserving their efficiency, robustness, and interpretability?}


To this end, we propose {\modelname}, a novel decision forest framework that leverages a LLM as an \textit{offline model designer}, guiding model construction without participating in inference. Rather than using the LLM for inference, {\modelname} engages it once during training to guide both split selection and leaf prediction. Our {\modelname} is fast, interpretable, and entirely LLM-free at test time. At its core, {\modelname} rethinks decision tree construction under few-shot supervision through a new framework for splitting and prediction. Our approach is two-fold.



First, we introduce \textit{semi-supervised} semantic tree induction, a split selection strategy that replaces conventional label-driven heuristics with LLM-guided semantic evaluation. Each candidate split is summarized into a prompt that captures labeled and unlabeled feature distributions along with local tree context. The LLM then ranks these candidates based on their semantic plausibility, effectively leveraging unlabeled data to guide generalization under limited supervision. Second, we propose \textit{in-context} leaf label inference, a one-time procedure for producing stable, static predictions at leaf nodes. Each decision path is translated into a natural language rule, which, together with its supporting examples, is used to prompt the LLM for a final prediction. This decouples inference from sparse empirical estimates, reducing variance while preserving efficiency. {\modelname} bakes in the structural bias of decision forests with the semantic abstraction capabilities of LLMs, without requiring fine-tuning or inference time access (Table~\ref{tab:llm_comparison}). It operates entirely in a few-shot, in-context regime and supports both classification and regression tasks. By repositioning the LLM as an offline architect rather than an online oracle, {\modelname} enables scalable, low-resource learning from tabular data.

We evaluate {\modelname} across a broad set of few-shot tabular benchmarks and find that it consistently matches or outperforms strong baselines, often achieving state-of-the-art or second-best performance with as few as 4 to 48 labeled examples. Notably, it retains strong performance on datasets that were unseen during LLM pretraining, underscoring its generalization ability beyond memorized priors.

\textbf{Our key contributions are as follows:}
\begin{itemize}
    \item We introduce the concept of using a LLM as an \textit{offline model designer}, guiding training-time structure induction without relying on the LLM at test time.
    \item We propose a novel semantic splitting strategy that replaces statistical heuristics with LLM-evaluated prompts over labeled and unlabeled data.
    \item We develop a single-pass, in-context leaf prediction mechanism that produces deterministic, low-variance outputs without empirical averaging or LLM inference.
    \item We demonstrate that {\modelname}achieves state-of-the-art few-shot performance on a wide range of tabular classification and regression benchmarks, without fine-tuning or inference-time prompting.
\end{itemize}

\begin{figure*}[tb]
\centering
\includegraphics[width=1\linewidth]{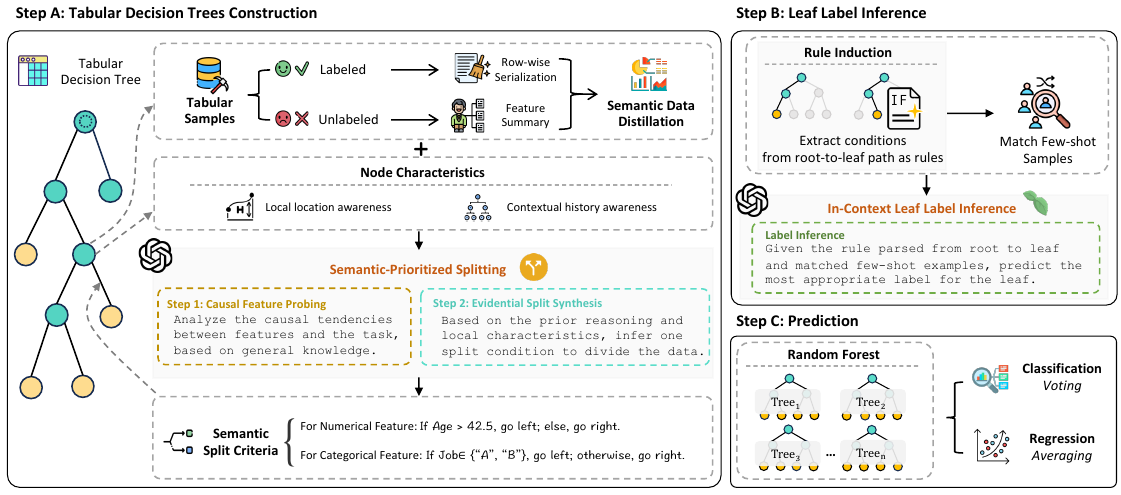}
\caption{An overview of our proposed {\modelname} framework for few-shot tabular learning.}
\label{fig:process}
\vspace{-1em}
\end{figure*}

\section{Related Work}

\paragraph{Few-Shot and Semi-Supervised Tabular Learning.}
Few-shot tabular learning poses a core challenge: enabling generalization from limited labeled data~\cite{chen2018closer, majumder2022few}. A common line of work addresses this by leveraging unlabeled data to learn expressive representations via semi- or self-supervised pretraining~\cite{somepalli2021saint, yoon2020vime, ucar2021subtab, bahri2022scarf, wang2022transtab}. In contrast, {\modelname} takes a fundamentally different approach: rather than using unlabeled data to pretrain encoders, it incorporates feature-level statistics from unlabeled examples directly into tree construction. Through \textit{semi-supervised semantic tree induction}, these statistics are summarized and passed to an LLM that guides split decisions based on semantic coherence, effectively treating unlabeled data as a prior over model structure, rather than input embeddings.

\paragraph{Decision Trees, Deep Models, and LLMs for Tabular Data.}
Modeling tabular data has traditionally relied on two dominant paradigms: tree-based ensembles such as GBDTs~\cite{chen2016xgboost, ke2017lightgbm}, and deep neural networks~\cite{kadra2021well, gorishniy2021revisiting, borisov2022deep}. While tree models encode strong inductive biases for hierarchical partitioning, their performance degrades in low-data regimes due to brittle splitting heuristics and high-variance leaf estimates. Deep models, in contrast, often fail to align with the structural properties of tabular data, resulting in poor generalization~\cite{mcelfresh2023neural, zabergja2024deep}.
{\modelname} addresses both limitations by replacing unstable statistical splits with LLM-guided semantic partitioning, and by stabilizing predictions through \textit{in-context leaf label inference}, a one-time procedure that bypasses noisy aggregation. In parallel, recent work has explored LLMs for tabular tasks~\cite{dinh2022lift, hegselmann2023tabllm}, leveraging fine-tuning~\cite{hu2021lora}, in-context prompting~\cite{nam2023semi}, serialization-based inference~\cite{nam2024tabular}, or synthetic data generation~\cite{han2024large}. However, these methods often incur high inference costs, lose structural information through serialization, or fail to produce standalone models~\cite{manikandan2023language}.
In contrast, {\modelname} casts the LLM as an \textit{offline model designer}, used solely during training to synthesize a semantically grounded decision forest. The resulting model preserves the computational efficiency of classical trees while benefiting from the semantic priors encoded in LLMs, without introducing inference-time latency or dependence.

\section{Method}

\subsection{Problem Formulation and Our Approach}

We consider the problem of few-shot tabular learning, where the goal is to learn a function $h: \mathbb{R}^d \to \mathcal{Y}$ from a small labeled dataset $\mathcal{D}_L = \{(x_i, y_i)\}_{i=1}^{N_L}$ and an optional unlabeled dataset $\mathcal{D}_U = \{x_j\}_{j=1}^{N_U}$, with $N_L \ll N_U$. Here, the labeled sample size $N_L$ corresponds to the \textit{shot count} in our experiments. Classical decision tree methods optimize each node split $s^*$ by optimizing statistical objectives such as information gain or Gini impurity. These heuristics, however, rely entirely on $\mathcal{D}_L$ and often fail in low-data regimes due to high variance and weak statistical signals. In contrast, {\modelname} constructs a forest $\mathcal{F} = \{\mathcal{T}_k\}_{k=1}^{K}$ of semantically grounded decision trees, where each split is selected via LLM-guided reasoning that integrates both labeled and unlabeled data.


At each internal node, we generate a semantically enriched prompt using natural language representations of local samples (via \textit{semantic data distillation}) and structural priors (node position and contextual history). The LLM evaluates optimal candidate splits: $s^* = \arg\max_{s \in \mathcal{S}} \Phi_{\mathcal{L}}(s \mid \mathcal{D}_L, \mathcal{D}_U, c)$, where $\Phi_{\mathcal{L}}$ denotes a plausibility score based on causal reasoning and $c$ encodes node-local features.
At the leaf level, label prediction is performed via in-context prompting: each decision path $r_\ell$ is translated into a rule $\Pi_{NL}(r_\ell)$ and a set of few-shot exemplars $\mathcal{D}_L^{\text{few}}$ is retrieved. The LLM then produces a static label: $\hat{y}_\ell = \arg\max_{y \in \mathcal{Y}} P_{\mathcal{L}}(y \mid \Pi_{NL}(r_\ell), \mathcal{D}_L^{\text{few}})$. At inference time, predictions from each tree are aggregated via majority voting (classification) or averaging (regression):
$\hat{y}(x') = \text{AGG}(\{\mathcal{T}_k(x')\}_{k=1}^{K})$. By incorporating \textit{unlabeled data} and \textit{structural context} during tree construction, and offloading \textit{label inference} to a rule-driven reasoning step, {\modelname} frames the LLM as a few-shot structure optimizer and label reasoner, used only during training.


\subsection{Semi-Supervised Semantic Tree Induction}

We propose \textit{semantic-prioritized splitting}, a novel decision tree induction framework that leverages LLMs as a symbolic reasoner to guide recursive partitioning. Unlike classical approaches that rely solely on label-based heuristics (e.g., information gain), our method incorporates both labeled and unlabeled data, making it particularly suited for few-shot regimes. The LLM uses sparse labeled examples for supervision and dense unlabeled instances to infer semantically coherent splits aligned with the underlying data manifold.

At each node $n$, given data $\mathcal{D}_n = \mathcal{D}_{L,n} \cup \mathcal{D}_{U,n}$, we perform \textit{semantic data distillation} to construct a prompt for the LLM.  Labeled portion $\mathcal{D}_{L,n}$ is converted into textual exemplars via \textit{row-wise serialization}, while unlabeled portion $\mathcal{D}_{U,n}$ are summarized into a compact representation of the marginal feature distribution $P(X)$ (the \textit{Feature Summary}). These components are merged with structural metadata, including the node’s depth and its contextual path from the root, to form the final prompt $\Pi_{\text{prompt}} = \Psi(\mathcal{D}_{L,n}, \mathcal{D}_{U,n}) \oplus \Gamma(p_n)$. Here, $\Psi$ is the distillation function, and $\Gamma(p_n)$ encodes the \textit{node characteristics} that provide (1) \textit{local depth awareness}, which calibrates split complexity, and (2) \textit{path history awareness}, which delineates the submanifold being partitioned.

The LLM processes this prompt in two steps: (1) \textit{Causal feature probing}, which hypothesizes relationships of the form $X_i \rightarrow Y$, and (2) \textit{Evidential split synthesis} which generates candidate split rules $s \in \mathcal{S}_n$. The final split is selected by maximizing a composite score:
\begin{equation}
\label{eq:semisupervised_split}
s_n^* = \underset{s \in \mathcal{S}_n}{\arg\max} \left[ \underbrace{\phi_{CP}^{\mathcal{L}}(s \mid \Pi_{\text{prompt}})}_{\text{Causal Plausibility}} \cdot \underbrace{\phi_{ES}^{\mathcal{L}}(s \mid \Pi_{\text{prompt}})}_{\text{Evidential Support}} \right]
\end{equation}
Here, $\phi_{CP}^{\mathcal{L}}$ measures semantic coherence with respect to a hypothesized causal relation, and $\phi_{ES}^{\mathcal{L}}$ reflects compatibility with evidence in the prompt.

\paragraph{Theoretical Analysis.}
Classical splitting criteria, such as impurity reduction $\Delta\mathcal{I}(s)$, rely exclusively on labeled data $\mathcal{D}_{L,n}$:
\begin{equation}
\label{eq:impurity_reduction}
\Delta\mathcal{I}(s) = \mathcal{I}(\mathcal{D}_{L,n}) - \sum_{k \in \{L, R\}} \frac{|\mathcal{D}_{L,n}^{(k)}|}{|\mathcal{D}_{L,n}|} \mathcal{I}(\mathcal{D}_{L,n}^{(k)})
\end{equation}
Because $\Delta\mathcal{I}(s)$ is computed solely from the empirical distribution $\hat{P}(Y \mid X)$ over $\mathcal{D}_{L,n}$, unlabeled data $\mathcal{D}_{U,n}$ is completely ignored: $\Delta\mathcal{I}(s; \mathcal{D}_{L,n} \cup \mathcal{D}_{U,n}) = \Delta\mathcal{I}(s; \mathcal{D}_{L,n})$. 
This fundamental limitation prevents classical trees from exploiting structural information in $\mathcal{D}_{L,n}$.
{\modelname} addresses this limitation by introducing a \textit{feature summary} as a proxy for the marginal distribution $P(X)$, estimated from $\hat{p}(x \mid \mathcal{D}_n)$. This enables the LLM to reason over the data geometry and apply the \textit{cluster assumption}, favoring splits $B_s$ that occur in low-density regions, i.e., minimizing $\int_{x \in B_s} \hat{p}(x \mid \mathcal{D}_n) \, dx$. By aligning splits with the intrinsic structure of the data, {\modelname} regularizes tree construction, mitigates overfitting to sparse labels, and improves generalization in few-shot regimes.


\subsection{In-Context Leaf Label Inference}

Once a decision tree is constructed, assigning a label to a test instance reaching a leaf node $\ell$ requires more than conventional majority voting, which is often unreliable in few-shot or low-resource regimes. We propose \textit{in-context leaf label inference}, a framework that reuses a pretrained LLM as a structured reasoning engine. Instead of relying on empirical aggregation over sparse leaf data, we prompt the LLM with a symbolic description of the decision path and a small set of semantically relevant exemplars, enabling context-aware prediction beyond the training distribution.

For each leaf $\ell$, we extract its decision path $r_\ell$ from the root and translate it into a natural language rule $\mathcal{R}\ell = \Pi{\text{NL}}(r_\ell)$ via \textit{Rule Induction} function. This rule captures the logic of the path as a structured, interpretable condition. To ground the inference, we retrieve a support set $\mathcal{D}L^{\text{few}} \subset \mathcal{D}L$ using a semantic retrieval function $\sigma$, which selects labeled examples most aligned with the semantics of $r\ell$:
$\mathcal{D}L^{\text{few}} = \sigma(\mathcal{D}L, r\ell)$.
We then compose a prompt using a template $\Pi{\text{ICL}}(\mathcal{R}\ell, \mathcal{D}_L^{\text{few}})$, which is passed to the LLM to generate a label via analogical reasoning and rule-based abstraction.
We formalize this as probabilistic inference over a latent space of reasoning traces $\mathcal{Z}$, where each $z \in \mathcal{Z}$ represents a candidate thought process induced by the LLM:
\begin{equation}
\hat{y}_\ell = \underset{y \in \mathcal{Y}}{\arg\max} \int_{\mathcal{Z}} P_{\mathcal{L}}(y \mid z, \mathcal{R}_\ell) \, P_{\mathcal{L}}(z \mid \text{prompt}_\ell) \, dz
\end{equation}
Here, $P{\mathcal{L}}(z \mid \text{prompt}\ell)$ captures the distribution over latent reasoning chains, and $P{\mathcal{L}}(y \mid z, \mathcal{R}_\ell)$ scores the plausibility of a predicted label given the rule and reasoning trace.

This formulation casts the LLM as a soft neuro-symbolic reasoner, integrating logical structure with contextual evidence to yield data-efficient predictions. By abstracting the decision path and grounding it in retrieved examples, our approach addresses key limitations of prior tree-based methods and opens new directions for LLM-driven inference in structured prediction.

\subsection{Overall Objective and Forest-Based Inference}

{\modelname} defines a semantically grounded random forest $\mathcal{F} = \{\mathcal{T}_k\}_{k=1}^{K}$, where each tree $\mathcal{T}_k$ is constructed via recursive application of the language-informed splitting objective (Eq.~\ref{eq:semisupervised_split}).
To promote diversity, each tree is trained on a bootstrap sample $\mathcal{D}^{(k)} \subset \mathcal{D}_L \cup \mathcal{D}_U$, and splits are selected from a random subset of features. This design preserves the statistical benefits of bagging while aligning tree structure with semantic priors induced by the LLM.

Inference on a test input $x'$ proceeds in two stages. First, each tree $\mathcal{T}_k$ routes $x'$ to a leaf node $\ell_k$, where a prediction $\hat{y}_k$ is produced via \textit{in-context leaf label inference}, incorporating both the decision rule and retrieved exemplars for contextual reasoning. Second, the predictions $\{\hat{y}_1, \ldots, \hat{y}_K\}$ are aggregated into the final output $\hat{y}(x’)$ using a task-specific rule: majority vote for classification or arithmetic mean for regression.
By integrating semantic tree construction with offline LLM-guided label reasoning, {\modelname} extends classical forests to the few-shot regime, achieving robust, data-efficient prediction without requiring inference-time access to the LLM.




\begin{table*}[!htb]
      \centering
     \caption{Classification performance (AUC) comparison of various models on 13 datasets under different few-shot settings.}
     \vspace{-0.5em}
      \renewcommand{\arraystretch}{0.8}
   \small
      \begin{adjustbox}{max width=\textwidth}
    \begin{tabular}{l|c|cccccccccccccc}
    \toprule
    \textbf{Data}  & \textbf{Shot}  & \textbf{CART}  & \multicolumn{1}{c}{\textbf{RF}} & \textbf{XGBoost} & \textbf{LogReg} & \textbf{SCARF} & \textbf{STUNT} & \textbf{TabPFN} & \textbf{TP-BERTa} & \textbf{TabLLM} & \textbf{TABLET} & \multicolumn{1}{c}{\textbf{LIFT}} & \textbf{FeatLLM} & \textbf{P2T}   & \textbf{Ours} \\
    \midrule
    
    \rowcolor{rowgray}
    \multirow{3}[2]{*}{adult} & 4     & 0.600  & 0.629  & 0.500  & 0.539  & 0.595  & 0.503  & 0.691  & 0.582  & 0.800  & 0.807  & 0.710  & \textbf{0.870 } & 0.765  & \underline{0.833 } \\
    \rowcolor{rowgray}
         adult & 8     & 0.618  & 0.666  & 0.596  & 0.603  & 0.658  & 0.532  & 0.733  & 0.610  & 0.795  & 0.826  & 0.694  & \textbf{0.880 } & 0.779  & \underline{0.827 } \\
    \rowcolor{rowgray}
          & 16    & 0.639  & 0.725  & 0.685  & 0.667  & 0.658  & 0.541  & 0.746  & 0.658  & 0.821  & 0.819  & 0.667  & \textbf{0.877 } & 0.777  & \underline{0.823 } \\
    \midrule
    
    \multirow{3}[2]{*}{blood} & 4     & 0.493  & 0.500  & 0.500  & 0.523  & 0.502  & 0.562  & 0.563  & 0.446  & 0.561  & \underline{0.626 } & 0.511  & 0.530  & 0.568  & \textbf{0.691 } \\
          & 8     & 0.542  & 0.575  & 0.567  & 0.564  & 0.617  & 0.553  & 0.604  & 0.473  & 0.578  & \underline{0.648 } & 0.551  & 0.598  & 0.618  & \textbf{0.698 } \\
          & 16    & 0.602  & 0.635  & 0.601  & 0.673  & 0.593  & 0.551  & 0.636  & 0.497  & 0.605  & 0.655  & 0.523  & 0.621  & \underline{0.664 } & \textbf{0.709 } \\
    \midrule

    \rowcolor{rowgray}
    \multirow{3}[2]{*}{bank} & 4     & 0.535  & 0.613  & 0.500  & 0.585  & 0.556  & 0.531  & 0.602  & 0.401  & 0.609  & \underline{0.829 } & 0.616  & 0.723  & 0.683  & \textbf{0.837 } \\
    \rowcolor{rowgray}
     bank     & 8     & 0.608  & 0.699  & 0.587  & 0.676  & 0.572  & 0.544  & 0.706  & 0.408  & 0.638  & \underline{0.830 } & 0.628  & 0.742  & 0.723  & \textbf{0.835 } \\
    \rowcolor{rowgray}
          & 16    & 0.642  & 0.730  & 0.675  & 0.738  & 0.584  & 0.559  & 0.773  & 0.447  & 0.661  & \underline{0.836 } & 0.616  & 0.752  & 0.740  & \textbf{0.843 } \\
    \midrule
    
    \multirow{3}[2]{*}{car} & 4     & 0.562  & 0.554  & 0.500  & 0.580  & 0.571  & 0.564  & 0.612  & 0.585  & 0.558  & 0.840  & 0.799  & 0.703  & 0.656  & \textbf{0.864 } \\
          & 8     & 0.581  & 0.643  & 0.620  & 0.622  & 0.622  & 0.609  & 0.673  & 0.660  & 0.617  & \textbf{0.918 } & 0.779  & 0.702  & 0.704  & \underline{0.879 } \\
          & 16    & 0.669  & 0.736  & 0.679  & 0.642  & 0.673  & 0.652  & 0.785  & 0.724  & 0.628  & 0.845  & 0.804  & 0.744  & 0.760  & \textbf{0.881 } \\
    \midrule

    \rowcolor{rowgray}
    \multirow{3}[2]{*}{communities} & 4     & 0.568  & 0.606  & 0.500  & 0.559  & 0.654  & 0.560  & 0.687  & 0.469  & N/A   & \textbf{0.799 } & 0.686  & 0.633  & 0.586  & \underline{0.784 } \\
    \rowcolor{rowgray}
     communities     & 8     & 0.603  & 0.718  & 0.664  & 0.586  & 0.726  & 0.612  & 0.743  & 0.519  & N/A   & \textbf{0.798 } & 0.725  & 0.691  & 0.594  & \underline{0.782 } \\
    \rowcolor{rowgray}
          & 16    & 0.623  & 0.738  & 0.689  & 0.639  & 0.745  & 0.646  & 0.769  & 0.556  & N/A   & \textbf{0.794 } & 0.725  & 0.699  & 0.675  & \underline{0.789 } \\
    \midrule
    
    \multirow{3}[2]{*}{credit-g} & 4     & 0.515  & 0.526  & 0.500  & 0.521  & 0.539  & 0.508  & 0.532  & 0.467  & 0.580  & \underline{0.623 } & 0.506  & 0.498  & 0.538  & \textbf{0.669 } \\
          & 8     & 0.543  & 0.571  & 0.546  & 0.600  & 0.548  & 0.506  & 0.584  & 0.469  & 0.624  & \underline{0.627 } & 0.499  & 0.531  & 0.588  & \textbf{0.687 } \\
          & 16    & 0.572  & 0.599  & 0.561  & 0.582  & 0.558  & 0.527  & 0.645  & 0.477  & 0.658  & 0.632  & 0.513  & 0.548  & \underline{0.661 } & \textbf{0.693 } \\
    \midrule

    \rowcolor{rowgray}
    \multirow{3}[2]{*}{cdc diabetes} & 4     & 0.539  & 0.526  & 0.500  & 0.565  & 0.604  & 0.557  & 0.605  & 0.541  & 0.624  & \textbf{0.702 } & 0.655  & 0.604  & 0.570  & \underline{0.700 } \\
    \rowcolor{rowgray}
    cdc diabetes      & 8     & 0.551  & 0.605  & 0.578  & 0.559  & 0.622  & 0.581  & 0.613  & 0.578  & 0.638  & \textbf{0.703 } & 0.656  & 0.590  & 0.609  & \underline{0.699 } \\
    \rowcolor{rowgray}
          & 16    & 0.562  & 0.611  & 0.600  & 0.590  & 0.600  & 0.572  & 0.630  & 0.602  & 0.620  & \underline{0.701 } & 0.649  & 0.633  & 0.599  & \textbf{0.705 } \\
    \midrule
    
    \multirow{3}[2]{*}{heart} & 4     & 0.595  & 0.711  & 0.500  & 0.510  & 0.816  & 0.611  & 0.767  & 0.497  & 0.734  & 0.858  & 0.652  & \underline{0.870 } & 0.602  & \textbf{0.887 } \\
          & 8     & 0.639  & 0.721  & 0.594  & 0.659  & 0.867  & 0.594  & 0.828  & 0.521  & 0.809  & 0.858  & 0.563  & \underline{0.880 } & 0.763  & \textbf{0.896 } \\
          & 16    & 0.695  & 0.840  & 0.804  & 0.742  & \underline{0.891 } & 0.632  & \underline{0.880 } & 0.630  & 0.848  & 0.857  & 0.570  & 0.877  & 0.763  & \textbf{0.898 } \\
    \midrule

    \rowcolor{rowgray}
    \multirow{3}[2]{*}{myocardial} & 4     & 0.513  & 0.519  & 0.500  & 0.536  & 0.551  & 0.527  & 0.522  & 0.516  & N/A   & \underline{0.609 } & 0.503  & 0.568  & 0.580  & \textbf{0.653 } \\
    \rowcolor{rowgray}
    myocardial      & 8     & 0.512  & 0.537  & 0.543  & 0.521  & 0.513  & 0.528  & 0.573  & 0.538  & N/A   & \underline{0.616 } & 0.519  & 0.549  & 0.583  & \textbf{0.660 } \\
    \rowcolor{rowgray}
          & 16    & 0.547  & 0.592  & 0.569  & 0.537  & 0.519  & 0.538  & 0.608  & 0.587  & N/A   & \underline{0.632 } & 0.570  & 0.577  & 0.576  & \textbf{0.687 } \\
    \midrule
    
    \multirow{3}[2]{*}{breast-w} & 4     & 0.836  & 0.960  & 0.500  & 0.783  & 0.985  & 0.883  & 0.986  & 0.923  & 0.985  & 0.974  & 0.759  & \underline{0.986 } & 0.979  & \textbf{0.992 } \\
          & 8     & 0.856  & 0.974  & 0.832  & 0.887  & 0.983  & 0.927  & 0.985  & 0.926  & 0.983  & 0.978  & 0.768  & \underline{0.987 } & 0.981  & \textbf{0.993 } \\
          & 16    & 0.851  & 0.970  & 0.876  & 0.962  & 0.985  & 0.921  & 0.984  & 0.934  & 0.979  & 0.978  & 0.912  & \underline{0.988 } & 0.985  & \textbf{0.993 } \\
    \midrule

    \rowcolor{rowgray}
    \multirow{3}[2]{*}{cultivars} & 4     & 0.514  & 0.493  & 0.500  & 0.504  & 0.494  & 0.514  & 0.499  & 0.445  & 0.524  & 0.525  & 0.531  & 0.528  & \underline{0.536 } & \textbf{0.661 } \\
    \rowcolor{rowgray}
     cultivars     & 8     & 0.486  & 0.497  & 0.503  & 0.528  & 0.519  & 0.498  & 0.541  & 0.495  & 0.554  & 0.546  & 0.522  & \underline{0.561 } & 0.543  & \textbf{0.653 } \\
    \rowcolor{rowgray}
          & 16    & 0.495  & 0.523  & 0.510  & 0.530  & 0.527  & 0.519  & 0.511  & 0.491  & \underline{0.584 } & 0.531  & 0.563  & 0.582  & 0.576  & \textbf{0.679 } \\
    \midrule
    
    \multirow{3}[2]{*}{NHANES} & 4     & 0.662  & 0.805  & 0.500  & 0.870  & 0.773  & 0.512  & 0.898  & 0.457  & \underline{0.975 } & 0.525  & 0.531  & 0.527  & 0.673  & \textbf{0.987 } \\
          & 8     & 0.962  & 0.900  & 0.880  & 0.968  & 0.812  & 0.518  & 0.976  & 0.519  & \textbf{0.999 } & 0.546  & 0.522  & 0.750  & 0.925  & \underline{0.998 } \\
          & 16    & 0.967  & 0.974  & 0.956  & 0.988  & 0.844  & 0.532  & \underline{0.999 } & 0.521  & 0.999  & 0.531  & 0.563  & 0.899  & 0.969  & \textbf{0.999 } \\
    \midrule

    \rowcolor{rowgray}
    \multirow{3}[2]{*}{gallstone} & 4     & 0.559  & 0.564  & 0.500  & 0.493  & 0.505  & 0.509  & \underline{0.570 } & 0.440  & 0.465  & 0.565  & 0.533  & 0.540  & 0.471  & \textbf{0.658 } \\
    \rowcolor{rowgray}
    gallstone      & 8     & 0.555  & 0.571  & 0.508  & 0.532  & 0.522  & 0.505  & \underline{0.597 } & 0.432  & 0.473  & 0.558  & 0.543  & 0.532  & 0.436  & \textbf{0.653 } \\
    \rowcolor{rowgray}
          & 16    & 0.592  & 0.591  & 0.589  & 0.596  & 0.537  & 0.528  & \underline{0.612 } & 0.426  & 0.461  & 0.563  & 0.529  & 0.557  & 0.481  & \textbf{0.662 } \\
    \midrule
    \multirow{3}[2]{*}{\textbf{Average}} & 4     & \cellcolor[rgb]{ .988,  .729,  .478} 0.576  & \cellcolor[rgb]{ .996,  .898,  .51} 0.616  & \cellcolor[rgb]{ .973,  .412,  .42} 0.500  & \cellcolor[rgb]{ .988,  .753,  .482} 0.582  & \cellcolor[rgb]{ .98,  .918,  .518} 0.627  & \cellcolor[rgb]{ .984,  .682,  .471} 0.565  & \cellcolor[rgb]{ .875,  .886,  .514} 0.656  & \cellcolor[rgb]{ .976,  .498,  .435} 0.521  & \cellcolor[rgb]{ .808,  .867,  .51} 0.674  & \cellcolor[rgb]{ .659,  .824,  .498} \underline{0.714 } & \cellcolor[rgb]{ .996,  .894,  .51} 0.615  & \cellcolor[rgb]{ .859,  .882,  .51} 0.660  & \cellcolor[rgb]{ .965,  .914,  .518} 0.631  & \cellcolor[rgb]{ .388,  .745,  .482} \textbf{0.786 } \\
          & 8     & \cellcolor[rgb]{ .988,  .725,  .478} 0.620  & \cellcolor[rgb]{ .984,  .918,  .518} 0.667  & \cellcolor[rgb]{ .988,  .71,  .475} 0.617  & \cellcolor[rgb]{ .992,  .808,  .494} 0.639  & \cellcolor[rgb]{ .996,  .902,  .514} 0.660  & \cellcolor[rgb]{ .976,  .529,  .439} 0.577  & \cellcolor[rgb]{ .804,  .867,  .51} 0.704  & \cellcolor[rgb]{ .973,  .412,  .42} 0.550  & \cellcolor[rgb]{ .82,  .871,  .51} 0.701  & \cellcolor[rgb]{ .694,  .835,  .502} \underline{0.727 } & \cellcolor[rgb]{ .984,  .694,  .471} 0.613  & \cellcolor[rgb]{ .863,  .882,  .51} 0.692  & \cellcolor[rgb]{ .922,  .902,  .514} 0.680  & \cellcolor[rgb]{ .388,  .745,  .482} \textbf{0.789 } \\
          & 16    & \cellcolor[rgb]{ .988,  .714,  .475} 0.650  & \cellcolor[rgb]{ .906,  .894,  .514} 0.713  & \cellcolor[rgb]{ .992,  .827,  .498} 0.676  & \cellcolor[rgb]{ .996,  .863,  .506} 0.684  & \cellcolor[rgb]{ .992,  .8,  .494} 0.670  & \cellcolor[rgb]{ .973,  .467,  .427} 0.594  & \cellcolor[rgb]{ .757,  .855,  .506} \underline{0.737 } & \cellcolor[rgb]{ .973,  .412,  .42} 0.581  & \cellcolor[rgb]{ .89,  .89,  .514} 0.715  & \cellcolor[rgb]{ .855,  .882,  .51} 0.721  & \cellcolor[rgb]{ .984,  .631,  .459} 0.631  & \cellcolor[rgb]{ .863,  .882,  .51} 0.720  & \cellcolor[rgb]{ .922,  .902,  .514} 0.710  & \cellcolor[rgb]{ .388,  .745,  .482} \textbf{0.797 } \\
    \bottomrule
        \end{tabular}%
        \end{adjustbox}
      \label{tab:cls_main}%
    \end{table*}%

\section{Experiments}

\textbf{Datasets.} 
We evaluate {\modelname} on a diverse suite of tabular datasets spanning both classification and regression tasks. For \textit{binary classification}, we include Adult~\citep{asuncion2007uci}, Bank~\citep{yeh2009knowledge}, Blood~\citep{moro2014data}, Credit-g~\citep{kadra2021well}, Cultivars~\citep{de2023dataset}, and NHANES\footnote{\url{archive.ics.uci.edu/dataset/887}}. For \textit{multi-class classification}, we use Car~\citep{kadra2021well}, CDC Diabetes\footnote{\url{kaggle.com/datasets/alexteboul/diabetes-health-indicators-dataset}}, Heart\footnote{\url{kaggle.com/datasets/fedesoriano/heart-failure-prediction}}, Communities~\citep{redmond2009communities}, and Myocardial~\citep{golovenkin2020myocardial}. For \textit{regression}, we evaluate on: 
Cultivars, Cpu\_small\footnote{\url{openml.org/search?type=data&id=562}}, Diamonds\footnote{\url{openml.org/search?type=data&id=42225}}, Plasma\_retinol\footnote{\url{openml.org/search?type=data&id=511}}, Forest-fires\footnote{\url{openml.org/search?type=data&id=42363}}, Housing\footnote{\url{openml.org/search?type=data&id=43996}}, Insurance\footnote{\url{kaggle.com/datasets/teertha/ushealthinsurancedataset}}, Bike~\citep{bike_sharing_275}, and Wine~\citep{cortez2009modeling}.

\textit{To reduce the risk of data contamination from the LLM’s pretraining corpus}~\citep{bordt2024elephants}, we additionally include two datasets that were publicly released after the GPT-4o model’s training cutoff (October 2023): Gallstone~\citep{esen2024early} for classification and Infrared\_Thermography\_Temperature~\citep{wang2023facial} for regression. Both datasets are sourced from the \textit{UCI Machine Learning Repository} and serve to evaluate post-training generalization on data that was inaccessible during LLM pretraining.

Our benchmark suite covers a broad range of application domains, including healthcare, finance, and food, and exhibits significant diversity in both feature dimensionality and dataset scale. For example, Myocardial includes up to 111 features, while the largest dataset contains over 253,680 samples. Full dataset statistics are provided in Appendix~A.



\begin{figure*}[!htb]
\centering
\includegraphics[width=1\linewidth]{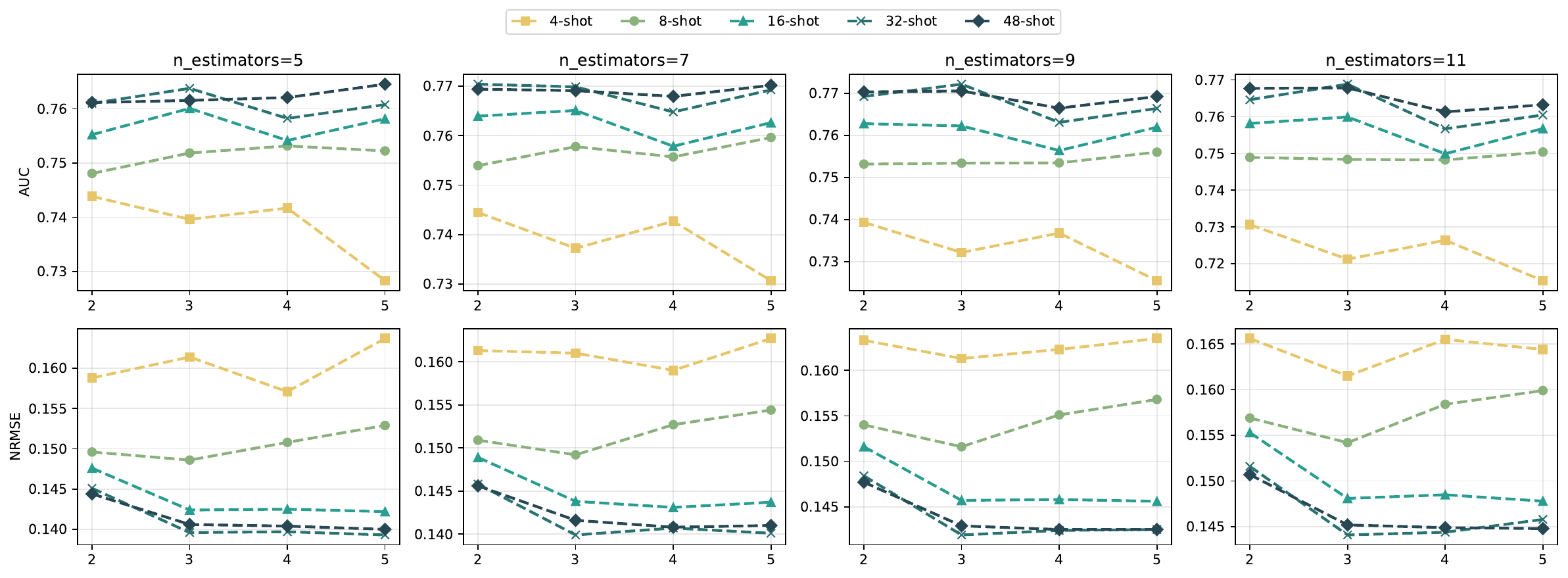}
\vspace{-1.5em}
\caption{Hyperparameter sensitivity analysis of {\modelname} with respect to tree depth and number of estimators under varying few-shot settings. The top row shows classification results, and the bottom row shows regression results.}
\label{fig:param}
\vspace{-1em}
\end{figure*}

\vspace{0.5em}
\noindent \textbf{Baselines and Implementation Details.}
To evaluate the effectiveness of our {\modelname}, we compare against a comprehensive set of baselines spanning both LLM-based and conventional non-LLM methods.

Among LLM-based approaches, we evaludate FeatLLM~\citep{han2024large}, LIFT~\citep{dinh2022lift}, TABLET~\citep{slack2023tablet}, TabLLM~\citep{hegselmann2023tabllm}, TP-BERTa~\citep{yan2024making}, and P2T~\citep{nam2024tabular}. \textsc{\modelname}, FeatLLM, LIFT, and TABLET use GPT-4o (gpt-4o-2024-11-20) as the base LLM, accessed via Azure. To eliminate output stochasticity, all LLM-based models are evaluated with temperature fixed to 0. Due to the high cost of fine-tuning and the small number of labeled examples, we use the in-context variant of LIFT only. For TabLLM and TP-BERTa, we adopt the model configurations and training protocols reported in their original papers, using T0 and RoBERTa as respective backbones.

For traditional baselines, we include CART~\citep{loh2011classification}, Random Forest (RF)\citep{breiman2001random}, XGBoost\citep{chen2016xgboost}, MLP~\citep{lecun2015deep}, ElasticNet~\citep{zou2005regularization}, and Logistic Regression (LogReg), along with specialized tabular learners such as TabPFN~\citep{hollmann2025tabpfn}, STUNT~\citep{nam2023stunt}, and SCARF~\citep{bahri2021scarf}.
Notably, FeatLLM, STUNT, SCARF, and \textsc{\modelname} leverage unlabeled data. For all such methods, unlabeled samples are consistently defined as the full dataset excluding the few-shot labeled examples and test set. All non-LLM baselines are tuned via grid search with k-fold cross-validation. Additional implementation details are provided in Appendix~B.

\begin{figure}[!htb]
\centering
\includegraphics[width=0.98\linewidth]{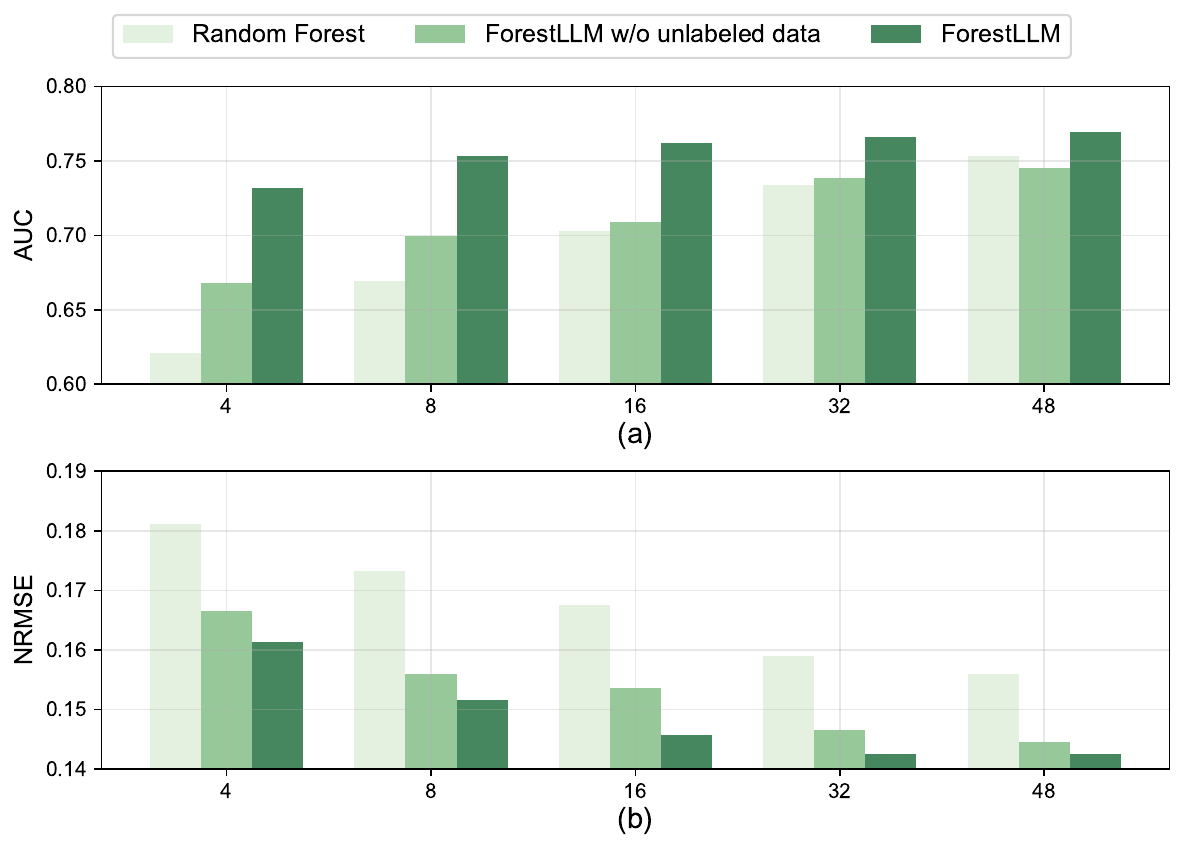}
\vspace{-1.0em}
\caption{Ablation study of the impact of unlabeled data. (a) Classification performance measured by AUC. (b) Regression performance measured by NRMSE.}
\label{fig:ablation}
\vspace{-1em}
\end{figure}

\subsection{Main Results}
Tables~\ref{tab:cls_main} and~\ref{tab:reg_main} present comprehensive performance comparisons across a wide range of classification and regression tasks. For each dataset, the best-performing method is shown in bold, and the second-best is \underline{underlined}. The bottom row summarizes the average performance across all datasets, visualized using a red–yellow–green gradient to indicate low, moderate, and high performance, respectively. 

To account for variance introduced by train/test split randomness in few-shot scenarios, we repeat each experiment 10 times using different random seeds (0–9) for data partitioning. We report the mean AUC (for classification) and NRMSE (for regression), with standard deviations 
provided in Appendix~E.

The results show that {\modelname} consistently achieves either the best or second-best performance across both classification and regression tasks. This superiority persists even in higher-shot settings (32- and 48-shot). Notably, {\modelname} also performs strongly on the Gallstone and Infrared\_Thermography\_Temperature datasets—both published after the LLM’s pretraining cutoff—demonstrating robust generalization to out-of-distribution and recently introduced data.

We further observe that LLM-based methods tend to outperform traditional machine learning approaches on classification tasks, while traditional methods often perform better on regression. 
This pattern is consistent with prior work highlighting the strength of LLMs in classification tasks~\citep{kostina2025large}, and their limitations in numerical reasoning and precise estimation~\citep{drinkall2024forecasting, siddiqui-etal-2025-evaluating}.
{\modelname} mitigates this limitation through a structured prediction mechanism based on random forest-style bagging and ensemble aggregation, rather than relying solely on raw LLM outputs. This design reduces uncertainty and mitigates errors introduced by the probabilistic nature of LLMs in numerical prediction tasks.

\begin{table*}[!t]
      \centering
      \caption{Regression performance (NRMSE) comparison of various models on 10 datasets under different few-shot settings. * refers to \texttt{infrared\_thermography\_temperature}.}
      \vspace{-0.5em}
      \renewcommand{\arraystretch}{0.35}
      \tiny
      \begin{adjustbox}{max width=\textwidth}
    \begin{tabular}{l|c|cccccccccc}
    \toprule
    \textbf{Data}  & \textbf{Shot}  & \textbf{CART}  & \textbf{RF} & \textbf{XGBoost} & \textbf{ElasticNet} & \textbf{MLP}   & \textbf{TabPFN} & \textbf{TP-BERTa} & \multicolumn{1}{c}{\textbf{LIFT}} & \textbf{P2T}   & \textbf{Ours} \\
    \midrule

    \rowcolor{rowgray}
    \multirow{3}[2]{*}{bike} & 4     & 0.220  & 0.194  & 0.224  & 0.266  & 0.225  & 0.192  & 0.264  & \underline{0.161 } & 0.190  & \textbf{0.156 } \\
    \rowcolor{rowgray}
    bike      & 8     & 0.300  & 0.197  & 0.243  & 0.211  & 0.247  & 0.191  & 0.262  & 0.259  & \underline{0.185 } & \textbf{0.149 } \\
    \rowcolor{rowgray}
          & 16    & 0.304  & 0.200  & 0.193  & 0.202  & 0.256  & 0.165  & 0.261  & \underline{0.156 } & 0.194  & \textbf{0.144 } \\
    \midrule
    
    \multirow{3}[2]{*}{cpu\_small} & 4     & 0.168  & 0.180  & 0.198  & 0.322  & 49.042  & \underline{0.167 } & 0.789  & 0.173  & 0.234  & \textbf{0.161 } \\
          & 8     & 0.172  & 0.167  & 0.187  & 0.389  & 9.865  & \underline{0.164 } & 0.770  & 0.172  & 0.167  & \textbf{0.154 } \\
          & 16    & 0.156  & 0.122  & 0.180  & 0.335  & 0.987  & 0.118  & 0.745  & \textbf{0.101 } & 0.147  & \underline{0.114 } \\
    \midrule

    \rowcolor{rowgray}
    \multirow{3}[2]{*}{diamonds} & 4     & 0.152  & 0.152  & 0.156  & 0.142  & 0.153  & 0.157  & 0.302  & \textbf{0.080 } & 0.116  & \underline{0.090 } \\
    \rowcolor{rowgray}
    diamonds      & 8     & 0.150  & 0.118  & 0.129  & 0.169  & 0.116  & 0.153  & 0.302  & \textbf{0.080 } & 0.109  & \underline{0.095 } \\
    \rowcolor{rowgray}
          & 16    & 0.126  & 0.120  & 0.134  & 0.123  & 0.115  & 0.119  & 0.302  & \textbf{0.082 } & 0.109  & \underline{0.095 } \\
    \midrule

    \multirow{3}[2]{*}{forest-fires} & 4     & 0.127  & 0.127  & 0.127  & 0.128  & 0.130  & \underline{0.127 } & 0.128  & 0.127  & 0.127  & \textbf{0.126 } \\
    & 8     & 0.204  & 0.143  & 0.140  & 0.170  & 0.128  & 0.170  & \underline{0.127 } & 0.153  & 0.136  & \textbf{0.126 } \\
    
          & 16    & 0.214  & 0.146  & 0.144  & 0.147  & 0.133  & 0.136  & \underline{0.126 } & 0.158  & 0.143  & \textbf{0.123 } \\
    \midrule

    \rowcolor{rowgray}
    \multirow{3}[2]{*}{houses} & 4     & 0.252  & 0.180  & \underline{0.156 } & 0.393  & 6.056  & 0.176  & 1.270  & 0.214  & 0.358  & \textbf{0.150 } \\
    \rowcolor{rowgray}
    houses      & 8     & 0.181  & 0.170  & 0.161  & 0.286  & 6.218  & \underline{0.142 } & 0.801  & 0.169  & 0.192  & \textbf{0.131 } \\
    \rowcolor{rowgray}
          & 16    & 0.175  & 0.150  & 0.156  & 0.183  & 2.840  & \underline{0.131 } & 0.285  & 0.148  & 0.163  & \textbf{0.128 } \\
    \midrule

    \multirow{3}[2]{*}{insurance} & 4     & 0.282  & 0.236  & 0.236  & 0.240  & 0.235  & 0.194  & 0.325  & \underline{0.125 } & 0.186  & \textbf{0.115 } \\
    
          & 8     & 0.295  & 0.255  & 0.245  & 0.224  & 0.207  & 0.163  & 0.325  & \underline{0.116 } & 0.145  & \textbf{0.107 } \\
    
          & 16    & 0.230  & 0.232  & 0.234  & 0.185  & 0.193  & 0.158  & 0.325  & \underline{0.116 } & 0.119  & \textbf{0.105 } \\
    \midrule

    \rowcolor{rowgray}
    \multicolumn{1}{l|}{\multirow{3}[2]{*}{plasma\_\newline{}retinol}} & 4     & 0.292  & 0.239  & \textbf{0.215 } & 0.490  & 0.396  & 0.227  & 0.633  & 0.331  & 0.412  & \underline{0.216 } \\
    \rowcolor{rowgray}
    retinol      & 8     & 0.304  & 0.240  & \underline{0.234 } & 0.388  & 0.472  & 0.242  & 0.631  & 0.340  & 0.316  & \textbf{0.203 } \\
    \rowcolor{rowgray}
          & 16    & 0.296  & 0.217  & 0.220  & 0.301  & 0.445  & \underline{0.203 } & 0.628  & 0.340  & 0.268  & \textbf{0.192 } \\
    \midrule

    \multirow{3}[2]{*}{wine} & 4     & 0.209  & 0.170  & 0.169  & 0.206  & 0.361  & 0.166  & \underline{0.150 } & 0.164  & 0.252  & \textbf{0.141 } \\
    
          & 8     & 0.188  & 0.167  & 0.161  & 0.193  & 0.228  & 0.153  & \underline{0.146 } & 0.158  & 0.183  & \textbf{0.137 } \\
    
          & 16    & 0.187  & 0.156  & 0.158  & 0.178  & 0.207  & 0.155  & \underline{0.148 } & 0.158  & 0.161  & \textbf{0.132 } \\
    \midrule

    \rowcolor{rowgray}
    \multirow{3}[2]{*}{cultivars} & 4     & 0.268  & 0.234  & \underline{0.218 } & 0.325  & 0.314  & 0.300  & 1.403  & 0.370  & 0.328  & \textbf{0.206 } \\
    \rowcolor{rowgray}
    cultivars      & 8     & 0.263  & 0.235  & 0.224  & 0.369  & 0.330  & \underline{0.221 } & 1.402  & 0.321  & 0.288  & \textbf{0.196 } \\
    \rowcolor{rowgray}
          & 16    & 0.277  & 0.234  & 0.231  & 0.224  & 0.369  & \underline{0.217 } & 1.400  & 0.270  & 0.266  & \textbf{0.198 } \\
    \midrule

    \multirow{3}[2]{*}{infrared\_t\_t*} & 4     & 0.130  & 0.116  & 0.151  & 0.152  & 0.288  & 0.138  & 6.628  & \underline{0.129 } & 1.637  & \textbf{0.108 } \\
    
         & 8     & 0.129  & 0.109  & 0.160  & 0.132  & 0.138  & 0.110  & 5.898  & \underline{0.105 } & 0.722  & \textbf{0.102 } \\
    
          & 16    & 0.117  & \underline{0.099 } & 0.109  & 0.120  & 0.140  & 0.112  & 5.036  & \textbf{0.098 } & 0.106  & 0.103  \\
    \midrule
    \multirow{3}[2]{*}{\textbf{Average}} & 4     & \cellcolor[rgb]{ 1,  .922,  .518} 0.210  & \cellcolor[rgb]{ .816,  .867,  .506} \underline{0.183 } & \cellcolor[rgb]{ .839,  .875,  .506} 0.185  & \cellcolor[rgb]{ 1,  .918,  .518} 0.266  & \cellcolor[rgb]{ .973,  .412,  .42} 5.720  & \cellcolor[rgb]{ .827,  .871,  .506} 0.184  & \cellcolor[rgb]{ .996,  .831,  .502} 1.189  & \cellcolor[rgb]{ .863,  .878,  .506} 0.187  & \cellcolor[rgb]{ 1,  .906,  .518} 0.384  & \cellcolor[rgb]{ .388,  .745,  .482} \textbf{0.147 } \\
          & 8     & \cellcolor[rgb]{ 1,  .918,  .518} 0.219  & \cellcolor[rgb]{ .773,  .855,  .502} 0.180  & \cellcolor[rgb]{ .847,  .878,  .506} 0.188  & \cellcolor[rgb]{ 1,  .906,  .518} 0.253  & \cellcolor[rgb]{ .973,  .412,  .42} 1.795  & \cellcolor[rgb]{ .686,  .827,  .498} \underline{0.171 } & \cellcolor[rgb]{ .988,  .647,  .467} 1.066  & \cellcolor[rgb]{ .839,  .875,  .506} 0.187  & \cellcolor[rgb]{ 1,  .91,  .518} 0.244  & \cellcolor[rgb]{ .388,  .745,  .482} \textbf{0.140 } \\
          & 16    & \cellcolor[rgb]{ 1,  .898,  .514} 0.208  & \cellcolor[rgb]{ .937,  .902,  .514} 0.168  & \cellcolor[rgb]{ 1,  .922,  .518} 0.176  & \cellcolor[rgb]{ 1,  .906,  .518} 0.200  & \cellcolor[rgb]{ .988,  .655,  .467} 0.569  & \cellcolor[rgb]{ .671,  .824,  .498} \underline{0.151 } & \cellcolor[rgb]{ .973,  .412,  .42} 0.926  & \cellcolor[rgb]{ .859,  .878,  .506} 0.163  & \cellcolor[rgb]{ .937,  .902,  .514} 0.168  & \cellcolor[rgb]{ .388,  .745,  .482} \textbf{0.133 } \\
        \bottomrule
        \end{tabular}%
        \end{adjustbox}
      \label{tab:reg_main}%
    \end{table*}%

\subsection{Ablations}
\noindent \textbf{Unlabeled Data Helps.}
We perform an ablation study to assess the contribution of unlabeled data to {\modelname}’s performance. As shown in Figure~\ref{fig:ablation}, we compare {\modelname} against (i) a variant trained using labeled data only, and (ii) a traditional Random Forest, under both classification and regression tasks. To ensure fair comparison, we adapt tree depth to the number of labeled examples. Specifically, for 4-, 8-, and 16-shot settings, we use \texttt{max\_depth=3} and \texttt{n\_estimators=9}; for 32- and 48-shot settings, we increase the depth to \texttt{max\_depth=5}, keeping \texttt{n\_estimators=9}. 
This adjustment accounts for the fact that, under extreme data scarcity, the limited number of labeled examples constrains tree growth depth, and shallow structures also help mitigate potential overfitting risks. In contrast, {\modelname} is capable of leveraging unlabeled data to construct deeper, semantically guided trees even in low-shot regimes, rendering it less sensitive to the number of labeled instances. The results show that {\modelname} consistently outperforms both baselines across all settings, confirming the effectiveness of utilizing unlabeled data for improved generalization.


\vspace{0.5em}
\noindent \textbf{Hyperparameters.}
We also evaluate the robustness of {\modelname} with respect to its key hyperparameters. As shown in Figure~\ref{fig:param}, we conduct a grid search over \texttt{max\_depth} and \texttt{n\_estimators}, reporting both AUC and NRMSE across 4-, 8-, 16-, 32-, and 48-shot settings. We vary \texttt{max\_depth} from 2 to 5 and sweep \texttt{n\_estimators} over {5, 7, 9, 11}. Results indicate that {\modelname} maintains strong performance across a wide range of configurations. While deeper trees and larger ensembles can offer marginal gains, the model is notably stable, especially in higher-shot settings (e.g., 32- and 48-shot). This robustness suggests that {\modelname} performs reliably without extensive hyperparameter tuning, making it a practical choice for real-world few-shot tabular learning.

\begin{figure}[!htb]
\centering
\includegraphics[width=0.98\linewidth]{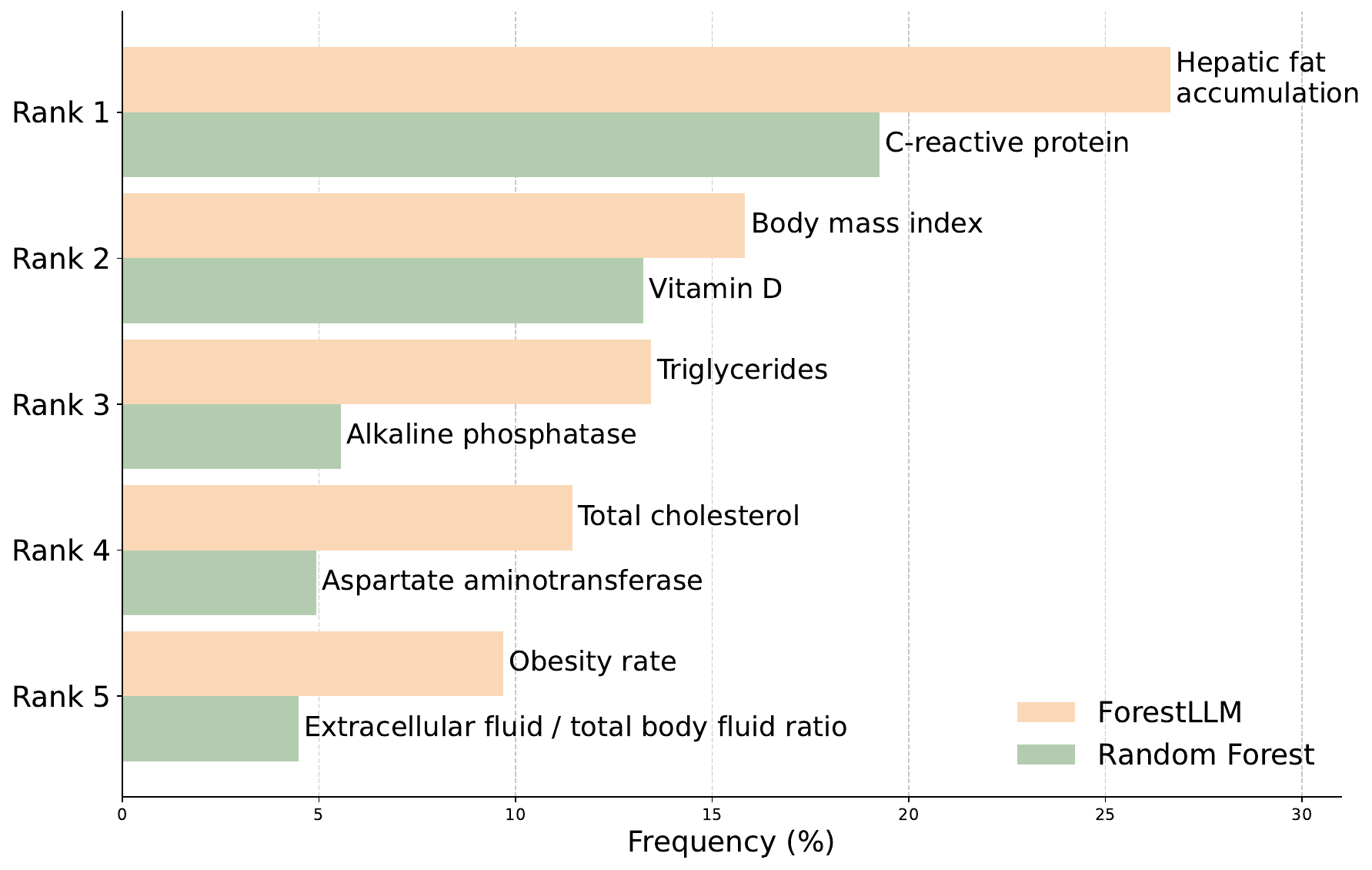}
\vspace{-1em}
\caption{Comparison of the top 5 most frequently split features on the Gallstone dataset.}
\label{fig:feature_comparison}
\vspace{-1em}
\end{figure}


\vspace{0.5em}
\noindent \textbf{Semantic Reasoning vs. Statistical Splitting.}
To assess interpretability in feature selection, we compared {\modelname} (under a 4-shot setup) with a fully supervised Random Forest on the Gallstone dataset. Feature importance was quantified via split frequency, revealing distinct selection patterns between the two models. {\modelname} prioritized upstream, causally relevant factors, such as hepatic fat, BMI, and cholesterol, consistent with known metabolic pathways implicated in gallstone formation~\citep{portincasa2023metabolic, wu2024association, tang2025association, deng2025relative}. In contrast, Random Forest emphasized downstream biomarkers including CRP, ALP, and AST, which are more reflective of inflammation and hepatic injury~\citep{zhou2021prediction, esen2024early, van2025putative}.

This divergence illustrates two fundamentally different paradigms: {\modelname} leverages semantic priors and causal abstractions derived from the LLM to surface clinically meaningful features, even in sparse data regimes. Random Forest, by contrast, relies on purely statistical associations, which often highlight features correlated with disease severity rather than etiology~\cite{esen2024early, van2025putative}. These results suggest that {\modelname} provides a causally grounded and interpretable alternative to conventional correlation-driven models, particularly in few-shot settings. Prior studies have highlighted the value of causal feature selection for improving interpretability and clinical insight (Chen, Zhang, and Qin 2022; Malec et al. 2023).

\section{Conclusion}

In this paper, we introduced {{\modelname}}, a novel framework that repositions LLMs as \textit{offline model designers}, bridging symbolic reasoning with structured inductive bias for few-shot tabular learning. By harnessing unlabeled data and LLM-driven semantic evaluations for both split selection and leaf-level label synthesis, {\modelname} constructs decision trees that retain the efficiency, interpretability, and inference-time frugality of classical forests, while achieving strong generalization in low-data regimes. Extensive experiments demonstrate that LLMs, when deployed as one-time architectural guides rather than inference-time oracles, can distill transferable knowledge into lightweight, symbolic models. Our work opens a new direction for leveraging foundation models as design-time optimizers for neuro-symbolic learners.


\clearpage


\appendix

\bibliography{aaai2026}


\clearpage
\onecolumn
\section{Appendix}

\subsection{A. Dataset Details}
We evaluate our method on a diverse suite of tabular datasets drawn from popular benchmarks: OpenML-CC18~\citep{oml-benchmarking-suites} for classification, OpenML-CTR23~\citep{fischer2023openml} for regression, and additional datasets from the UCI Machine Learning Repository and Kaggle. Table~\ref{tab:datasets} provides a summary organized by task type: binary classification, multi-class classification, and regression. For each dataset, we report key statistics, including the number of samples, number of features, label distribution, and dataset identifier. These datasets span a wide range of domains, scales, and structural properties, providing a rigorous testbed for assessing generalization and robustness in few-shot tabular learning.


\begin{table*}[htbp]
  \centering
  \caption{Dataset Characteristics}
    \begin{tabular}{lccccc}
    \toprule
    Dataset & \#Samples  & \#Features & \#Label ratio(\%) & Source & ID/Name \\
    \midrule
    \rowcolor[rgb]{ .749,  .749,  .749} \multicolumn{6}{c}{Binary Classification} \\
    \midrule
    adult & 48842 & 14    & 76:24 & OpenML & 1590 \\
    bank  & 45211 & 16    & 88:12 & OpenML & 1461 \\
    blood & 748   & 4     & 76:24 & OpenML & 1464 \\
    breast-w & 699   & 9     & 66:34 & OpenML & 15 \\
    credit-g & 1000  & 20    & 70:30 & OpenML & 31 \\
    cultivars & 320   & 10    & 50:50 & UCI   & \multicolumn{1}{p{15.165em}}{\makecell[c]{Forty Soybean Cultivars \\ from Subsequent Harvests}} \\
    gallstone & 320   & 38    & 50:50 & UCI   & Gallstone \\
    NHANES & 6287  & 8     & 84:16 & UCI   & \multicolumn{1}{p{15.165em}}{\makecell[c]{National Health and Nutrition Health \\ Survey 2013-2014 Age Prediction Subset}} \\
    \midrule
    \rowcolor[rgb]{ .749,  .749,  .749} \multicolumn{6}{c}{Multi-class Classification} \\
    \midrule
    car   & 1728  & 6     & 70:22:4:4 & OpenML & 40975 \\
    communities & 1994  & 103   & 34:33:33 & UCI   & Communities and Crime \\
    myocardial & 1700  & 111   & 22:78 & UCI   & Myocardial infarction complications \\
    heart & 918   & 11    & 45:55 & Kaggle & Heart Failure Prediction Dataset \\
    cdc diabetes & 253680 & 21    & 84:14:2 & Kaggle & Diabetes Health Indicators Dataset \\
    \midrule
    \rowcolor[rgb]{ .749,  .749,  .749} \multicolumn{6}{c}{Regression} \\
    \midrule
    Cpu\_small & 8192  & 12    & N/A   & OpenML & 562 \\
    diamonds & 53940 & 19    & N/A   & OpenML & 42225 \\
    forest-fires & 517   & 13    & N/A   & OpenML & 42363 \\
    housing & 20640 & 9     & N/A   & OpenML & 43996 \\
    plasma\_retinol & 315   & 13    & N/A   & OpenML & 511 \\
    bike  & 17389 & 11    & N/A   & UCI   & Bike Sharing \\
    cultivars & 320   & 10    & N/A   & UCI   & \multicolumn{1}{p{15.165em}}{\makecell[c]{Forty Soybean Cultivars \\ from Subsequent Harvests}} \\
    \makecell[l]{infrared\_thermography\_\\temperature} & 1020  & 33    & N/A   & UCI   & Infrared Thermography Temperature \\
    wine  & 4898  & 10    & N/A   & UCI   & Wine Quality \\
    insurance & 1338  & 7     & N/A   & Kaggle & US Health Insurance Dataset \\
    \bottomrule
    \bottomrule
    \end{tabular}%
  \label{tab:datasets}%
\end{table*}%

\subsection{B. Implementation Details}
\noindent\textbf{Data Preparation.} 
Following the popular setting~\citep{han2024large}, we partition each dataset into training and test sets, reserving 20\% of the data for testing. For classification tasks, we apply stratified sampling to preserve the label distribution across classes. From the training set, we construct a $k$-shot subset for supervision, selecting examples in a class-balanced manner whenever possible. In scenarios with limited data or many classes, we permit approximate balancing to maintain representativeness. For regression tasks, we discretize the continuous response variable into quantile-based bins and sample within each bin to maintain the marginal target distribution.
To account for variability due to stochastic sampling, especially prominent in few-shot regimes, we repeat each experiment using 10 different random seeds (0 through 9) and report mean performance metrics. All methods receive the same train/test partitions under each seed to ensure consistency and fairness.
To minimize confounding effects introduced by preprocessing, we adopt a lightweight and uniform imputation strategy. For LLM-based methods, all missing values—whether numerical or categorical—are replaced with the string ``\texttt{Unknown}'', which serves as a symbolic placeholder compatible with text prompts. For classical models, we impute categorical variables with ``\texttt{Unknown}'' and continuous variables with the column-wise mean. This design ensures that preprocessing does not bias the evaluation of semantic modeling capabilities or statistical performance across methods.

\vspace{0.5em}
\noindent\textbf{Hyperparameter Tuning.} 
Hyper-parameter selection is essential for fair comparison, particularly for classical machine learning baselines such as Logistic Regression (LogReg), Random Forest (RF), XGBoost, ElasticNet, Multilayer Perceptron (MLP), and TabPFN. We perform hyperparameter tuning for these models using grid search combined with $k$-shot cross-validation on the training data. Specific hyperparameter grids are detailed in Tables~\ref{tab:cart}--\ref{tab:tabpfn}.
For {\modelname}, we perform hyperparameter tuning via grid search without cross-validation to minimize API costs. In the few-shot regime, conventional train/validation splits (e.g., 70\%/30\%) yield validation sets too small for stable evaluation. Instead, we evaluate each configuration directly on the training set, which we find to be a practical compromise under severe data constraints. The hyperparameter search space is summarized in Table~\ref{tab:hyper_params_forestllm}.


\vspace{0.5em}
\noindent\textbf{Evaluation.}  
Classification performance is reported using the macro-averaged Area Under the ROC Curve (AUC), computed via the \texttt{roc\_auc\_score} function from \texttt{sklearn.metrics}. For LLM-based baselines including LIFT, TABLET, and P2T, prediction confidences are derived from token-level log-probabilities returned by the OpenAI GPT-4o API, following best practices for prompt-based classification. For regression, we report normalized root mean squared error (NRMSE), with normalization based on the standard deviation of the target variable in the test set.
All reported metrics are averaged over 10 independent trials with different train/test seeds. Experiments are executed on a dedicated Ubuntu 22.04.4 LTS server equipped with an AMD EPYC 7V13 64-core CPU, 432 GB of RAM, and two NVIDIA A100 80GB GPUs. OpenAI GPT-4o (engine: \texttt{gpt-4o-2024-11-20}) is accessed via the Azure OpenAI API, with temperature set to 0 for deterministic outputs across all LLM-based baselines. For ForestLLM, we follow this setting during leaf label assignment to ensure output consistency, while setting a higher temperature of 0.5 during tree construction to encourage diversity across trees.


\begin{table}[!htp]
  \centering
  \caption{Hyperparameter search space for CART}
    \begin{tabular}{ll}
    \toprule
    \multicolumn{1}{l|}{Parameter} & Search space \\
    \midrule
    criterion & \makecell[l]{gini, entropy\\squared\_error, friedman\_mse, absolute\_error} \\
    max\_depth & 3, 5, 7, 9, 11 \\
    \bottomrule
    \end{tabular}%
  \label{tab:cart}%
\end{table}%

\begin{table}[!htp]
  \centering
  \caption{Hyperparameter search space for Random Forest}
    \begin{tabular}{ll}
    \toprule
    \multicolumn{1}{l|}{Parameter} & Search space \\
    \midrule
    n\_estimators & 2, 4, 8, 16, 32, 64, 128, 256 \\
    bootstrap & True, False \\
    max\_depth & 3, 5, 7, 9, 11 \\
    max\_features & 0.5, 0.7, 1.0 \\
    \bottomrule
    \end{tabular}%
  \label{tab:rf}%
\end{table}%

\begin{table}[!htp]
  \centering
  \caption{Hyperparameter search space for XGBoost}
    \begin{tabular}{ll}
    \toprule
    \multicolumn{1}{l|}{Parameter} & Search space \\
    \midrule
    max\_depth & 3, 5, 7, 9, 11 \\
    alpha & 1e-8, 1e-7, 1e-6, 1e-5, 1e-4, 1e-3, 1e-2, 1e-1, 1, 10, 100 \\
    lambda & 1e-8, 1e-7, 1e-6, 1e-5, 1e-4, 1e-3, 1e-2, 1e-1, 1 \\
    eta   & 0.01, 0.03, 0.1, 0.3 \\
    \bottomrule
    \end{tabular}%
  \label{tab:xgboost}%
\end{table}%

\begin{table}[!htp]
  \centering
  \caption{Hyperparameter search space for LogReg}
    \begin{tabular}{ll}
    \toprule
    \multicolumn{1}{l|}{Parameter} & Search space \\
    \midrule
    C     & 1e-5, 1e-4, 1e-3, 1e-2, 1e-1, 1, 10, 100 \\
    penalty & l1, l2 \\
    \bottomrule
    \end{tabular}%
  \label{tab:logreg}%
\end{table}%

\begin{table}[!htp]
  \centering
  \caption{Hyperparameter search space for ElasticNet}
    \begin{tabular}{ll}
    \toprule
    \multicolumn{1}{l|}{Parameter} & Search space \\
    \midrule
    alpha & 1e-5, 1e-4, 1e-3, 1e-2, 1e-1, 1, 10, 100 \\
    l1\_ratio & 0.0,  1.0 \\
    \bottomrule
    \end{tabular}%
  \label{tab:elastic}%
\end{table}%

\begin{table}[!htp]
  \centering
  \caption{Hyperparameter search space for MLP}
    \begin{tabular}{ll}
    \toprule
    \multicolumn{1}{l|}{Parameter} & Search space \\
    \midrule
    hidden\_layer\_sizes & (64, ), (128,), (256,), (512,), (1024,) \\
    alpha & 0.1, 0.01, 0.001 \\
    learning\_rate\_init & 0.1, 0.01, 0.001 \\
    \bottomrule
    \end{tabular}%
  \label{tab:mlp}%
\end{table}%

\begin{table}[!htp]
  \centering
  \caption{Hyperparameter search space for TabPFN}
    \begin{tabular}{ll}
    \toprule
    \multicolumn{1}{l|}{Parameter} & Search space \\
    \midrule
    n\_estimators & 2, 4, 6, 8, 10, 12 \\
    \bottomrule
    \end{tabular}%
  \label{tab:tabpfn}%
\end{table}%

\begin{table}[!htp]
  \centering
  \caption{Hyperparameter search space for {\modelname}}
    \begin{tabular}{ll}
    \toprule
    \multicolumn{1}{l|}{Parameter} & Search space \\
    \midrule
    n\_estimators & 3, 4, 5, 6, 7, 8, 9, 10 \\
    max\_depth & 2, 3, 4, 5, 6, 7 \\
    bootstrap & True \\
    max\_features & 0.9 \\
    \bottomrule
    \end{tabular}%
  \label{tab:hyper_params_forestllm}%
\end{table}%

\subsection{C. Case Study of Reasoning During Tree Splits}
\begin{figure}[htbp]
\centering
\includegraphics[width=1\linewidth]{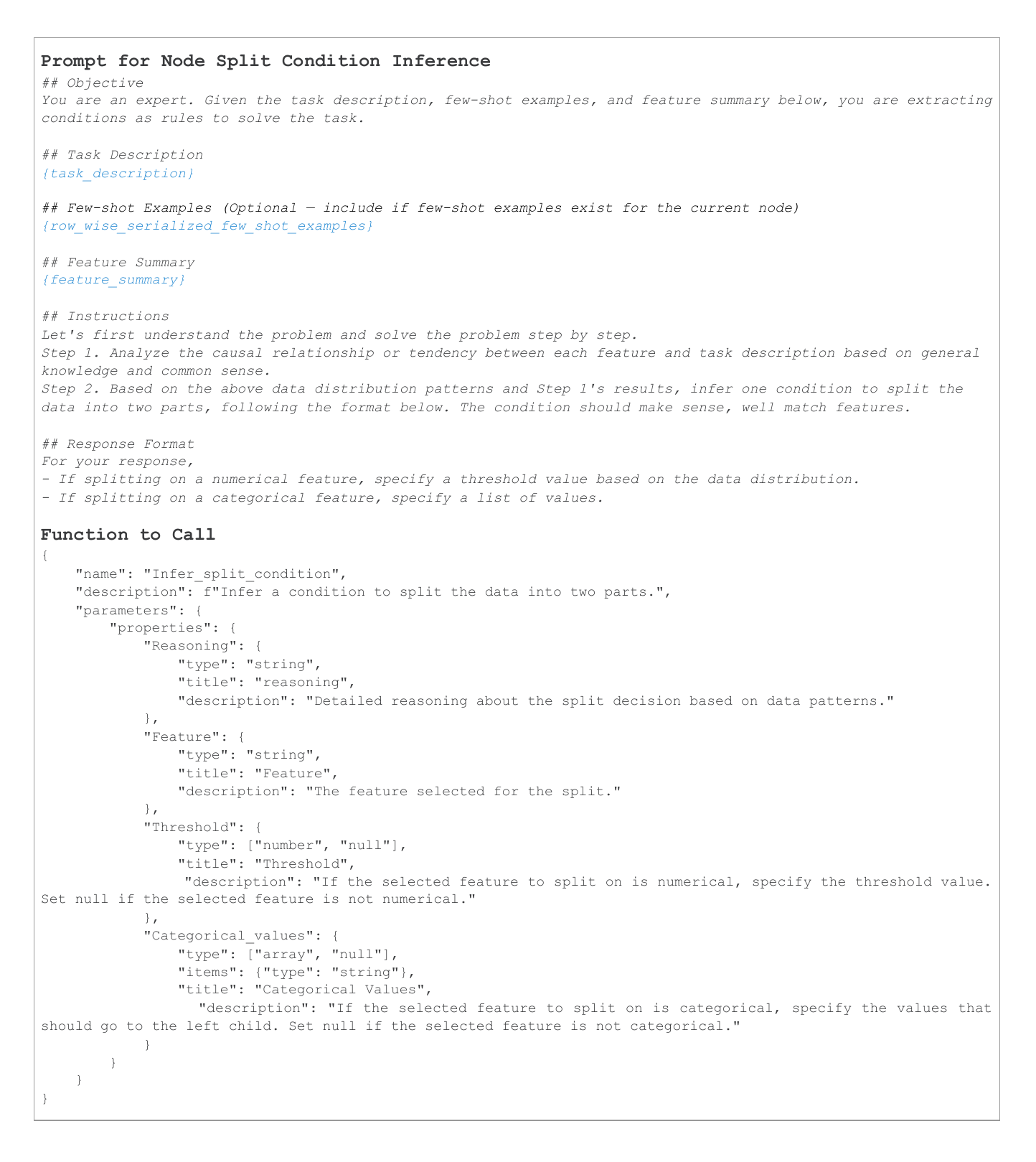}
\caption{
Prompt template and function calling schema used by {\modelname} to generate structured and interpretable split conditions. The \texttt{task\_description} provides a concise natural language description of the prediction task, drawn from domain knowledge or dataset metadata (e.g., “Given this person’s characteristics, does this person have diabetes?”). The \texttt{row\_wise\_serialized\_few\_shot\_examples} lists labeled samples under the current node in a row-wise serialized format, serving as few-shot examples to guide the model’s reasoning. The \texttt{feature\_summary} summarizes the features of all samples within the current node — including both labeled and unlabeled data — describing feature types, value distributions for categorical features, and statistics for numerical features.
}
\label{fig:prompt_tree}
\vspace{-0.em}
\end{figure}

\begin{figure}[htbp]
\centering
\includegraphics[width=1\linewidth]{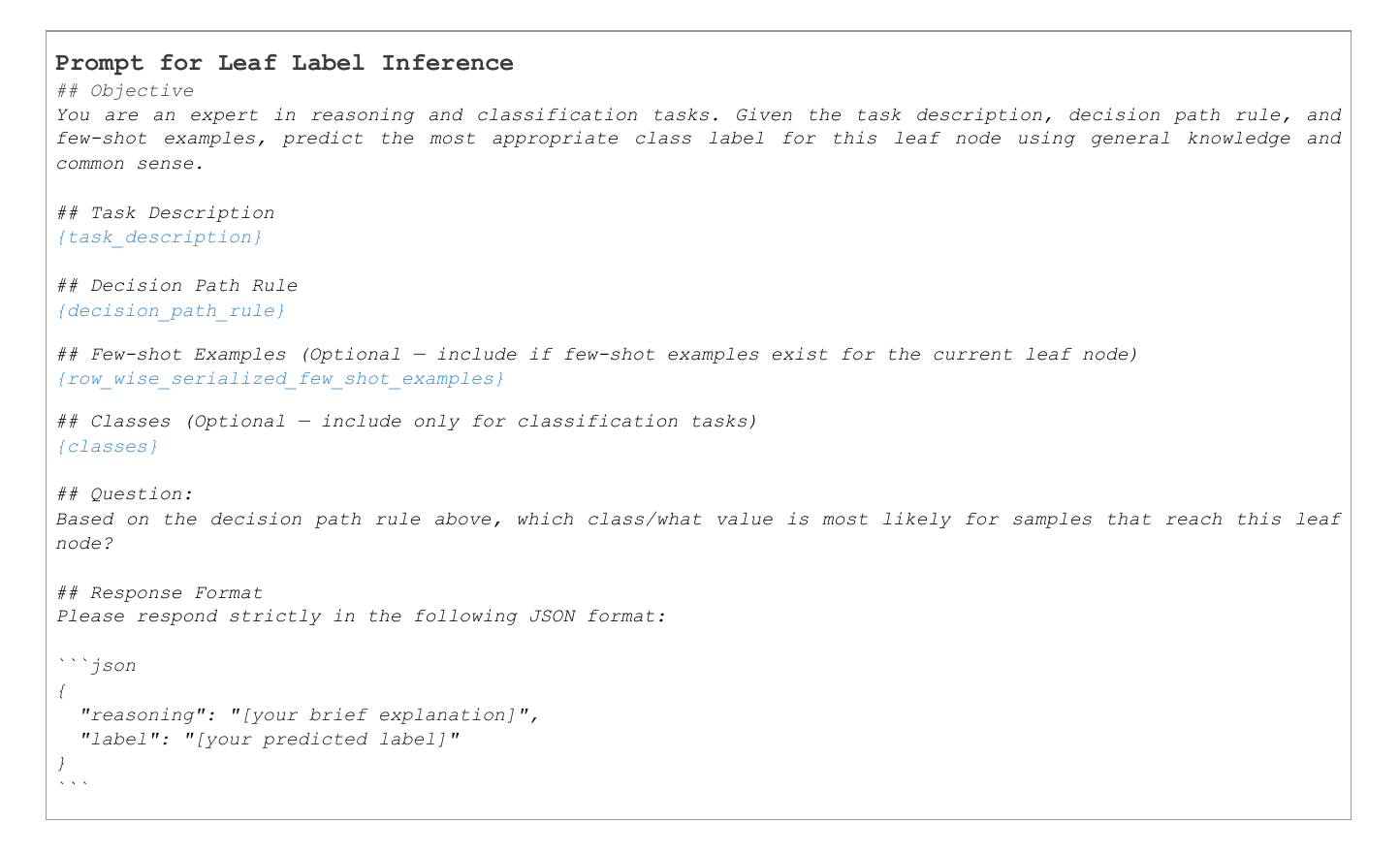}
\caption{
Prompt template used by {\modelname} for leaf node label inference.The \texttt{task\_description} provides a concise natural language description of the prediction task. The \texttt{decision\_path\_rule} summarizes the conditions along the path from the root to the current leaf node, representing the accumulated split decisions applied to reach this node. The \texttt{row\_wise\_serialized\_few\_shot\_examples} lists representative labeled samples that have arrived at the current leaf node, formatted in a row-wise serialized manner to support in-context few-shot reasoning. The \texttt{classes} enumerates all possible class labels for the classification task, including those that may not appear in the few-shot examples.
}
\label{fig:prompt_leaf}
\vspace{-0.em}
\end{figure}

\begin{figure}[htbp]
\centering
\includegraphics[width=1\linewidth]{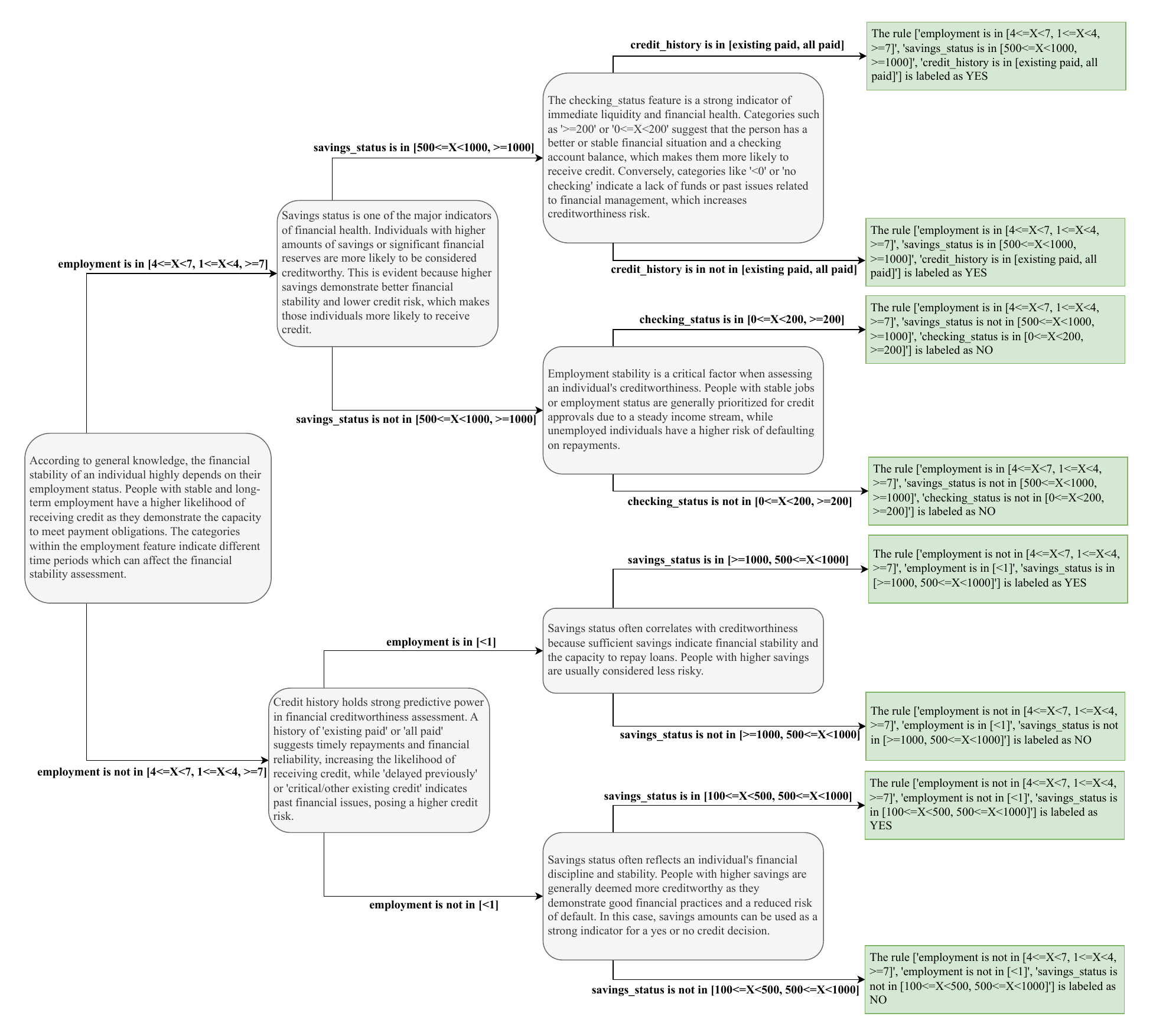}
\caption{A decision tree generated by {\modelname} on the Credit-g dataset.}
\label{fig:credit-g}
\vspace{-0.em}
\end{figure}

\begin{figure}[htbp]
\centering
\includegraphics[width=1\linewidth]{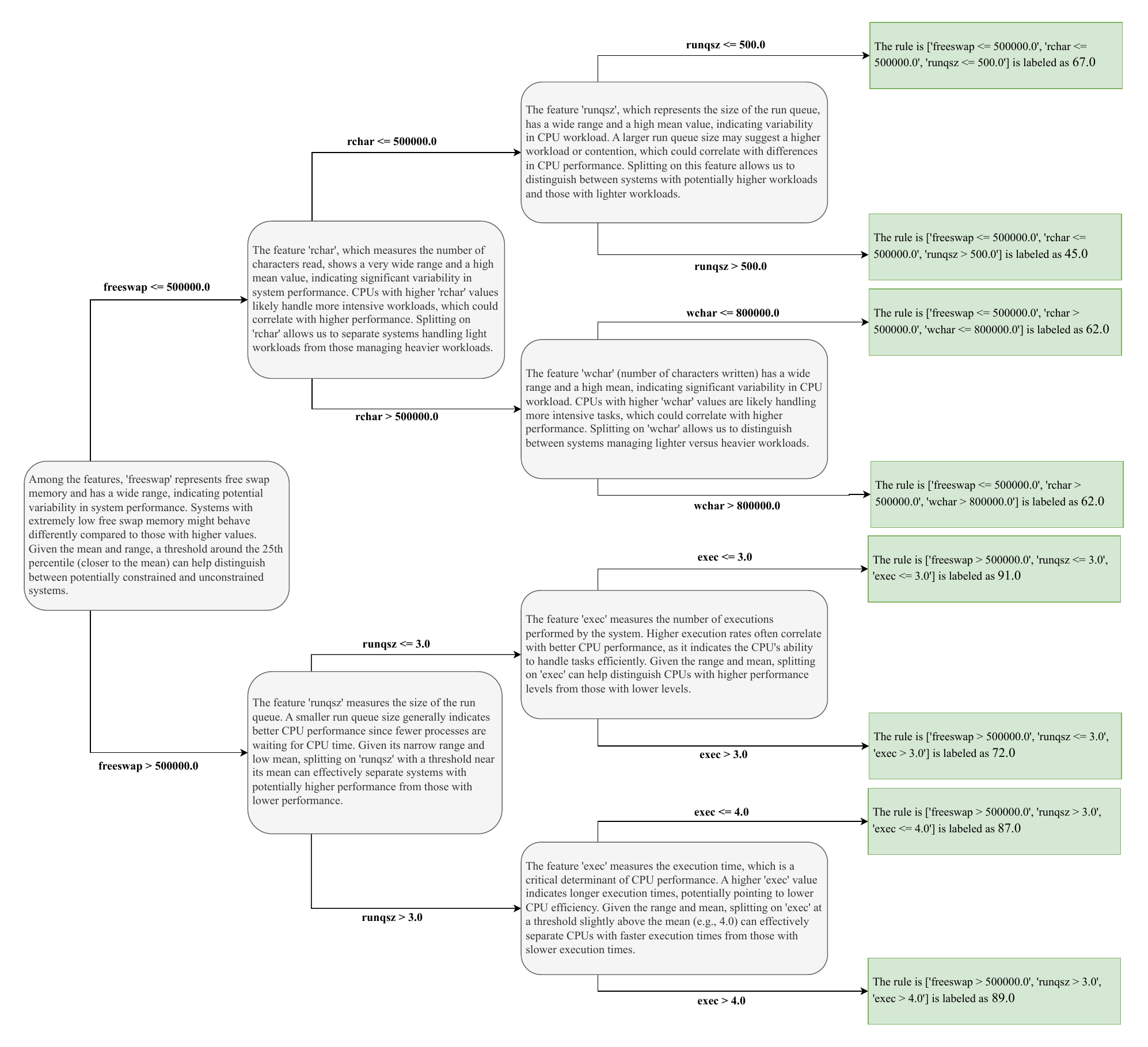}
\caption{A decision tree generated by {\modelname} on the Cpu\_small dataset.}
\label{fig:cpu_small}
\vspace{-0.em}
\end{figure}

To examine how {\modelname} performs semantic reasoning during decision tree construction, we visualize its generated reasoning traces, split criteria, and leaf label predictions for both classification and regression tasks (Figures~\ref{fig:credit-g} and~\ref{fig:cpu_small}). To ensure consistency and parseability in split decisions, we adopt structured prompting via function calling. The corresponding prompting template and function schema are shown in Figure~\ref{fig:prompt_tree}, while the leaf inference prompt is provided in Figure~\ref{fig:prompt_leaf}.


\vspace{0.5em}
\noindent We select two representative datasets under a 4-shot setting, one for classification and one for regression, to illustrate {\modelname}’s reasoning process:
\begin{itemize}
    \item \texttt{Credit-g} is a binary classification dataset containing 1,000 loan applicants, with approximately 700 labeled as ``good credit'' and 300 as ``bad credit.'' It consists of 20 features, including employment status, savings level, and credit history, and is widely used in credit risk prediction and interpretability research.
    
    \item \texttt{Cpu\_small} is a system performance regression dataset collected from a multi-user Sun Sparcstation system. Under real-world workloads, the system records performance metrics every 5 seconds, such as exec (number of system executions per second), rchar (characters read per second), runqsz (run queue size), and freemem (available memory pages). The task is to predict the CPU's user-mode utilization (usr) based on 12 continuous system-level features.
\end{itemize}

\vspace{0.5em}
\noindent By examining the reasoning traces generated during node splitting, we find that {\modelname} not only identifies predictive features but also provides semantically meaningful justifications grounded in commonsense and domain-specific knowledge. For example, in the \texttt{Credit-g} dataset, the model selects attributes such as employment status, savings level, and credit history, accompanied by explanations like “stable employment is a key factor in creditworthiness” and “higher savings indicate stronger repayment ability.” In the \texttt{Cpu\_small} dataset, it highlights features such as rchar, runqsz, and exec, correctly associating them with system throughput and scheduling behavior, indicating an understanding of low-level performance dynamics.


\vspace{0.5em}
\noindent Moreover, {\modelname} demonstrates the ability to infer semantically meaningful split thresholds by jointly leveraging the distributions of both labeled and unlabeled data. In \texttt{Cpu\_small}, for instance, the model selects a \texttt{freeswap} threshold of 500{,}000 to distinguish between resource-constrained and resource-rich systems; it splits on \texttt{exec} at 3.0 or 4.0 to identify differences in execution efficiency; and sets \texttt{runqsz} at 500.0 or 3.0 to reflect varying levels of system load. These thresholds are not only statistically valid but also align with intuitive interpretations of performance bottlenecks, highlighting {\modelname}’s capacity for human-aligned, data-driven reasoning under limited supervision.


\vspace{0.5em}
\noindent When assigning labels at leaf nodes, {\modelname} eschews conventional strategies such as majority voting or numerical averaging. Instead, it performs semantic generalization over the decision path by integrating few-shot labeled examples, domain knowledge, and logical reasoning. In the \texttt{Credit-g} dataset, for example, a path characterized by``stable employment + sufficient savings + good credit history'' is semantically interpreted as a low-risk credit profile, leading to the label ``YES.'' In \texttt{Cpu\_small}, a path reflecting ``ample system resources and efficient execution'' yields a high predicted utilization score. This path-aware reasoning framework allows {\modelname} to transcend low-level statistical heuristics, producing label assignments that are both interpretable and contextually aligned with the underlying semantics of the task.


\vspace{0.5em}
\noindent Overall, {\modelname} exhibits strong semantic reasoning and interpretability throughout the decision tree construction process. Operating under few-shot conditions with limited supervision, the model effectively fuses pretrained language knowledge with distributional cues from both labeled and unlabeled data to derive meaningful split criteria and label assignments. This synergy enables {\modelname} to overcome the inherent limitations of data scarcity, enhancing both its adaptability and the transparency of its predictions in low-resource tabular learning scenarios.


\subsection{D. Results with Different LLM Backbones}
To evaluate the impact of different LLM backbones on the performance of \textsc{ForestLLM}, we conduct additional experiments using \texttt{gpt-4-32k-0613} and \texttt{gpt-3.5-turbo-0125}, both deployed via Azure, as shown in Tables~\ref{tab:backbone_cls} and~\ref{tab:backbone_reg}. Although \texttt{GPT-4-32k} is based on an older architecture and lacks the enhanced capabilities of \texttt{GPT-4o}, it delivers comparable results across a wide range of datasets and even surpasses \texttt{GPT-4o} in certain cases. Interestingly, despite sharing the same knowledge cutoff date (September 2021), \texttt{GPT-4-32k} consistently outperforms \texttt{GPT-3.5-turbo} across tasks. This performance gap suggests that the underlying reasoning capabilities of the model, not simply knowledge recency, are critical for success in structured, few-shot tabular settings.


\begin{table*}[htbp]
  \centering
  \caption{Effect of LLM backbone choice on classification performance}
  \renewcommand{\arraystretch}{0.8}
    \begin{tabular}{l|c|ccc}
    \toprule
    Data  & Shot  & GPT-4o & GPT-4-32k & GPT-3.5-turbo \\
    \midrule
    \multirow{5}[2]{*}{adult} & 4     & 0.833 ± 0.041 & \textbf{0.838±0.030} & 0.816±0.034 \\
          & 8     & 0.827 ± 0.020 & \textbf{0.839±0.021} & 0.824±0.022 \\
          & 16    & 0.823 ± 0.028 & \textbf{0.835±0.027} & 0.823±0.014 \\
          & 32    & 0.846 ± 0.020 & \textbf{0.847±0.020} & 0.814±0.026 \\
          & 48    & 0.834 ± 0.032 & \textbf{0.852±0.018} & 0.833±0.011 \\
    \midrule
    \multirow{5}[2]{*}{blood} & 4     & 0.691 ± 0.066 & \textbf{0.719±0.060} & 0.699±0.051 \\
          & 8     & 0.698 ± 0.050 & 0.720±0.035 & \textbf{0.727±0.030} \\
          & 16    & 0.709 ± 0.042 & \textbf{0.734±0.038} & 0.729±0.038 \\
          & 32    & 0.723 ± 0.043 & \textbf{0.743±0.037} & 0.739±0.035 \\
          & 48    & 0.724 ± 0.049 & \textbf{0.744±0.030} & 0.732±0.035 \\
    \midrule
    \multirow{5}[2]{*}{bank} & 4     & 0.837 ± 0.014 & \textbf{0.843±0.009} & 0.755±0.036 \\
          & 8     & 0.835 ± 0.013 & \textbf{0.842±0.012} & 0.747±0.042 \\
          & 16    & \textbf{0.843 ± 0.008} & 0.843±0.012 & 0.758±0.024 \\
          & 32    & 0.841 ± 0.010 & \textbf{0.847±0.009} & 0.740±0.031 \\
          & 48    & \textbf{0.843 ± 0.011} & 0.841±0.011 & 0.749±0.036 \\
    \midrule
    \multirow{5}[2]{*}{car} & 4     & 0.864 ± 0.034 & \textbf{0.878±0.022} & 0.771±0.030 \\
          & 8     & 0.879 ± 0.028 & \textbf{0.903±0.018} & 0.728±0.022 \\
          & 16    & 0.881 ± 0.021 & \textbf{0.893±0.023} & 0.755±0.026 \\
          & 32    & 0.881 ± 0.029 & \textbf{0.909±0.021} & 0.769±0.030 \\
          & 48    & 0.893 ± 0.020 & \textbf{0.910±0.022} & 0.758±0.030 \\
    \midrule
    \multirow{5}[2]{*}{communities} & 4     & \textbf{0.784 ± 0.012} & 0.774±0.006 & 0.643±0.043 \\
          & 8     & \textbf{0.782 ± 0.008} & 0.771±0.008 & 0.627±0.037 \\
          & 16    & \textbf{0.789 ± 0.010} & 0.772±0.010 & 0.642±0.042 \\
          & 32    & \textbf{0.781 ± 0.013} & 0.780±0.010 & 0.660±0.026 \\
          & 48    & \textbf{0.779 ± 0.011} & 0.774±0.011 & 0.661±0.046 \\
    \midrule
    \multirow{5}[2]{*}{credit-g} & 4     & 0.669 ± 0.042 & \textbf{0.670±0.034} & 0.625±0.019 \\
          & 8     & \textbf{0.687 ± 0.053} & 0.686±0.050 & 0.628±0.030 \\
          & 16    & \textbf{0.693 ± 0.044} & 0.679±0.039 & 0.650±0.034 \\
          & 32    & \textbf{0.709 ± 0.031} & 0.679±0.034 & 0.630±0.034 \\
          & 48    & \textbf{0.713 ± 0.045} & 0.704±0.035 & 0.627±0.030 \\
    \midrule
    \multirow{5}[2]{*}{cdc diabetes} & 4     & 0.700 ± 0.012 & \textbf{0.702±0.010} & 0.612±0.030 \\
          & 8     & 0.699 ± 0.012 & \textbf{0.703±0.012} & 0.594±0.034 \\
          & 16    & \textbf{0.705 ± 0.011} & 0.704±0.011 & 0.593±0.035 \\
          & 32    & 0.709 ± 0.005 & \textbf{0.710±0.011} & 0.594±0.026 \\
          & 48    & 0.706 ± 0.009 & \textbf{0.706±0.007} & 0.598±0.019 \\
    \midrule
    \multirow{5}[2]{*}{heart} & 4     & 0.887 ± 0.022 & \textbf{0.894±0.015} & 0.834±0.038 \\
          & 8     & 0.896 ± 0.015 & \textbf{0.899±0.012} & 0.785±0.022 \\
          & 16    & \textbf{0.898 ± 0.011} & 0.893±0.017 & 0.773±0.024 \\
          & 32    & \textbf{0.896 ± 0.013} & 0.895±0.012 & 0.790±0.032 \\
          & 48    & \textbf{0.898 ± 0.018} & 0.891±0.014 & 0.794±0.022 \\
    \midrule
    \multirow{5}[2]{*}{myocardial} & 4     & 0.653 ± 0.048 & 0.619±0.069 & \textbf{0.667±0.037} \\
          & 8     & \textbf{0.660 ± 0.046} & 0.623±0.053 & 0.643±0.038 \\
          & 16    & \textbf{0.687 ± 0.048} & 0.645±0.025 & 0.647±0.041 \\
          & 32    & \textbf{0.681 ± 0.055} & 0.639±0.022 & 0.644±0.028 \\
          & 48    & \textbf{0.658 ± 0.034} & 0.656±0.023 & 0.636±0.027 \\
    \midrule
    \multirow{5}[2]{*}{breast-w} & 4     & 0.992 ± 0.006 & \textbf{0.993±0.005} & 0.989±0.008 \\
          & 8     & \textbf{0.993 ± 0.006} & 0.992±0.006 & 0.991±0.007 \\
          & 16    & 0.993 ± 0.006 & \textbf{0.994±0.005} & 0.992±0.005 \\
          & 32    & \textbf{0.994 ± 0.004} & 0.993±0.006 & 0.990±0.006 \\
          & 48    & \textbf{0.994 ± 0.006} & 0.992±0.006 & 0.992±0.006 \\
    \midrule
    \multirow{5}[2]{*}{cultivars} & 4     & \textbf{0.661 ± 0.031} & 0.653±0.044 & 0.630±0.040 \\
          & 8     & 0.653 ± 0.025 & \textbf{0.657±0.032} & 0.595±0.034 \\
          & 16    & \textbf{0.679 ± 0.027} & 0.677±0.044 & 0.628±0.030 \\
          & 32    & \textbf{0.676 ± 0.041} & 0.670±0.055 & 0.631±0.039 \\
          & 48    & 0.670 ± 0.049 & \textbf{0.671±0.045} & 0.639±0.034 \\
    \midrule
    \multirow{5}[2]{*}{NHANES} & 4     & \textbf{0.987 ± 0.010} & 0.974±0.055 & 0.892±0.128 \\
          & 8     & \textbf{0.998 ± 0.002} & 0.996±0.003 & 0.887±0.080 \\
          & 16    & \textbf{0.999 ± 0.001} & 0.996±0.006 & 0.947±0.049 \\
          & 32    & \textbf{0.999 ± 0.001} & 0.996±0.008 & 0.953±0.037 \\
          & 48    & \textbf{0.999 ± 0.001} & 0.995±0.007 & 0.964±0.054 \\
    \midrule
    \multirow{5}[2]{*}{gallstone} & 4     & 0.658 ± 0.042 & 0.674±0.058 & \textbf{0.698±0.037} \\
          & 8     & 0.653 ± 0.054 & 0.666±0.066 & \textbf{0.679±0.045} \\
          & 16    & 0.662 ± 0.045 & 0.674±0.042 & \textbf{0.677±0.058} \\
          & 32    & 0.659 ± 0.045 & 0.637±0.049 & \textbf{0.667±0.033} \\
          & 48    & 0.663 ± 0.054 & 0.656±0.046 & \textbf{0.689±0.038} \\
    \bottomrule
    \end{tabular}%
  \label{tab:backbone_cls}%
\end{table*}%

\begin{table*}[htbp]
  \centering
  \caption{Effect of LLM backbone choice on regression performance}
  \renewcommand{\arraystretch}{0.8}
    \begin{tabular}{l|c|ccc}
    \toprule
    Data  & Shot  & GPT-4o & GPT-4-32k &  GPT-3.5-turbo \\
    \midrule
    \multirow{5}[2]{*}{bike} & 4     & 0.156 ± 0.007 & \textbf{0.155±0.009} & 0.175±0.014 \\
          & 8     & 0.149 ± 0.007 & \textbf{0.145±0.006} & 0.177±0.012 \\
          & 16    & \textbf{0.144 ± 0.011} & 0.146±0.008 & 0.170±0.006 \\
          & 32    & 0.142 ± 0.006 & \textbf{0.142±0.004} & 0.168±0.007 \\
          & 48    & 0.141 ± 0.007 & \textbf{0.139±0.004} & 0.170±0.006 \\
    \midrule
    \multirow{5}[2]{*}{cpu\_small} & 4     & 0.161 ± 0.038 & \textbf{0.153±0.008} & 0.188±0.016 \\
          & 8     & 0.154 ± 0.024 & \textbf{0.151±0.026} & 0.195±0.010 \\
          & 16    & 0.114 ± 0.027 & \textbf{0.112±0.027} & 0.172±0.011 \\
          & 32    & 0.116 ± 0.022 & \textbf{0.115±0.021} & 0.176±0.009 \\
          & 48    & \textbf{0.122 ± 0.021} & 0.128±0.015 & 0.177±0.009 \\
    \midrule
    \multirow{5}[2]{*}{diamonds} & 4     & 0.090 ± 0.010 & \textbf{0.089±0.008} & 0.141±0.055 \\
          & 8     & 0.095 ± 0.013 & \textbf{0.092±0.012} & 0.158±0.033 \\
          & 16    & 0.095 ± 0.007 & \textbf{0.090±0.008} & 0.114±0.012 \\
          & 32    & \textbf{0.090 ± 0.005} & 0.092±0.004 & 0.117±0.014 \\
          & 48    & 0.090 ± 0.005 & \textbf{0.090±0.004} & 0.117±0.013 \\
    \midrule
    \multirow{5}[2]{*}{forest-fires} & 4     & 0.126 ± 0.029 & \textbf{0.125±0.029} & 0.126±0.029 \\
          & 8     & 0.126 ± 0.029 & 0.125±0.029 & \textbf{0.124±0.030} \\
          & 16    & 0.123 ± 0.029 & \textbf{0.122±0.029} & 0.123±0.027 \\
          & 32    & \textbf{0.122 ± 0.029} & 0.122±0.030 & 0.122±0.030 \\
          & 48    & \textbf{0.118 ± 0.031} & 0.118±0.031 & 0.121±0.029 \\
    \midrule
    \multirow{5}[2]{*}{houses} & 4     & \textbf{0.150 ± 0.030} & 0.151±0.028 & 0.180±0.056 \\
          & 8     & \textbf{0.131 ± 0.011} & 0.133±0.014 & 0.150±0.016 \\
          & 16    & \textbf{0.128 ± 0.011} & 0.128±0.011 & 0.143±0.013 \\
          & 32    & \textbf{0.127 ± 0.011} & 0.128±0.010 & 0.146±0.012 \\
          & 48    & \textbf{0.126 ± 0.011} & 0.127±0.009 & 0.148±0.011 \\
    \midrule
    \multirow{5}[2]{*}{insurance} & 4     & 0.115 ± 0.021 & \textbf{0.112±0.017} & 0.141±0.029 \\
          & 8     & \textbf{0.107 ± 0.014} & 0.107±0.015 & 0.133±0.023 \\
          & 16    & \textbf{0.105 ± 0.012} & 0.105±0.014 & 0.150±0.018 \\
          & 32    & \textbf{0.100 ± 0.010} & 0.100±0.012 & 0.153±0.022 \\
          & 48    & 0.102 ± 0.014 & \textbf{0.099±0.011} & 0.157±0.025 \\
    \midrule
    \multirow{5}[2]{*}{plasma\_retinol} & 4     & 0.216 ± 0.038 & \textbf{0.209±0.031} & 0.213±0.041 \\
          & 8     & 0.203 ± 0.028 & 0.203±0.034 & \textbf{0.201±0.027} \\
          & 16    & 0.192 ± 0.018 & \textbf{0.191±0.017} & 0.195±0.022 \\
          & 32    & \textbf{0.193 ± 0.017} & 0.193±0.021 & 0.198±0.026 \\
          & 48    & \textbf{0.193 ± 0.019} & 0.194±0.023 & 0.198±0.024 \\
    \midrule
    \multirow{5}[2]{*}{wine} & 4     & 0.141 ± 0.010 & 0.141±0.010 & \textbf{0.134±0.004} \\
          & 8     & 0.137 ± 0.006 & \textbf{0.135±0.006} & 0.136±0.006 \\
          & 16    & \textbf{0.132 ± 0.002} & 0.132±0.005 & 0.136±0.005 \\
          & 32    & 0.131 ± 0.002 & \textbf{0.130±0.003} & 0.138±0.005 \\
          & 48    & 0.131 ± 0.003 & \textbf{0.130±0.003} & 0.141±0.006 \\
    \midrule
    \multirow{5}[2]{*}{cultivars} & 4     & \textbf{0.206 ± 0.021} & 0.213±0.027 & 0.216±0.033 \\
          & 8     & \textbf{0.196 ± 0.017} & 0.199±0.017 & 0.204±0.017 \\
          & 16    & 0.198 ± 0.023 & 0.198±0.024 & \textbf{0.198±0.019} \\
          & 32    & \textbf{0.193 ± 0.019} & 0.194±0.019 & 0.196±0.017 \\
          & 48    & \textbf{0.191 ± 0.022} & 0.192±0.017 & 0.198±0.016 \\
    \midrule
    \multirow{5}[2]{*}{\makecell[l]{infrared\_thermography\_\\temperature}} & 4     & 0.108 ± 0.012 & \textbf{0.105±0.012} & 0.126±0.009 \\
          & 8     & 0.102 ± 0.010 & \textbf{0.100±0.010} & 0.135±0.026 \\
          & 16    & \textbf{0.103 ± 0.008} & 0.107±0.010 & 0.129±0.011 \\
          & 32    & \textbf{0.097 ± 0.007} & 0.100±0.010 & 0.125±0.010 \\
          & 48    & \textbf{0.096 ± 0.009} & 0.101±0.007 & 0.127±0.010 \\
    \bottomrule
    \end{tabular}%
  \label{tab:backbone_reg}%
\end{table*}%

\subsection{E. Full Results}
This section presents the complete experimental results across all datasets, baselines, and shot configurations (4, 8, 16, 32, and 48). While the main paper reports a representative subset due to space constraints, we include the full performance tables here for both classification and regression tasks, as shown in Tables~\ref{tab:cls_appendix} and~\ref{tab:reg_appendix}. These tables provide a comprehensive view of model behavior across varying supervision levels and further validate the robustness of our findings.



\begin{table*}[htbp]
  \centering
  \caption{Performance comparison of traditional methods and LLM-based methods across various datasets and 4-, 8-, 16-, 32-, and 48-shot settings for classification tasks. The best-performing method is bolded, and the second-best is underlined. Each value represents the AUC score, reported as the mean ± standard deviation across 10 random seeds.}
    \resizebox{\linewidth}{!}{%
    \begin{tabular}{l|c|cccccccccccccc}
    \toprule
    \multirow{3}[6]{*}{Dataset } & \multirow{3}[6]{*}{Shot} & \multicolumn{7}{c|}{Traditional Method}               & \multicolumn{7}{c}{LLM-based Method} \\
\cmidrule{3-16}          &       & \multicolumn{7}{c|}{Training Required}                & \multicolumn{2}{c|}{\makecell[c]{Fine-tuned \\ Required}} & \multicolumn{5}{c}{No Fine-tuning Required} \\
\cmidrule{3-16}          &       & CART  & \multicolumn{1}{c}{\makecell[c]{Random \\ Forest}} & XGBoost & LogReg & SCARF & STUNT & TabPFN & TP-BERTa & TabLLM & TABLET & LIFT  & FeatLLM & P2T   & Ours \\
    \midrule
    \multirow{5}[2]{*}{adult} & 4     & 0.600 ± 0.075 & 0.629 ± 0.131 & 0.500 ± 0.000 & 0.539 ± 0.057 & 0.595 ± 0.131 & 0.503 ± 0.030 & 0.691 ± 0.159 & 0.582 ± 0.096 & 0.800 ± 0.019 & 0.807 ± 0.039 & 0.710 ± 0.052 & \textbf{0.870 ± 0.025} & 0.765 ± 0.035 & \underline{0.833 ± 0.041} \\
          & 8     & 0.618 ± 0.091 & 0.666 ± 0.121 & 0.596 ± 0.081 & 0.603 ± 0.070 & 0.658 ± 0.124 & 0.532 ± 0.033 & 0.733 ± 0.132 & 0.610 ± 0.095 & 0.795 ± 0.050 & 0.826 ± 0.024 & 0.694 ± 0.039 & \textbf{0.880 ± 0.009} & 0.779 ± 0.036 & \underline{0.827 ± 0.020} \\
          & 16    & 0.639 ± 0.098 & 0.725 ± 0.092 & 0.685 ± 0.120 & 0.667 ± 0.066 & 0.658 ± 0.122 & 0.541 ± 0.025 & 0.746 ± 0.089 & 0.658 ± 0.056 & 0.821 ± 0.025 & 0.819 ± 0.020 & 0.667 ± 0.038 & \textbf{0.877 ± 0.013} & 0.777 ± 0.030 & \underline{0.823 ± 0.028} \\
          & 32    & 0.672 ± 0.060 & 0.759 ± 0.074 & 0.743 ± 0.089 & 0.687 ± 0.057 & 0.717 ± 0.097 & 0.532 ± 0.021 & 0.791 ± 0.089 & 0.659 ± 0.067 & 0.819 ± 0.029 & 0.820 ± 0.029 & 0.763 ± 0.057 & \textbf{0.874 ± 0.013} & 0.787 ± 0.041 & \underline{0.846 ± 0.020} \\
          & 48    & 0.672 ± 0.029 & 0.795 ± 0.051 & 0.757 ± 0.093 & 0.726 ± 0.076 & 0.763 ± 0.072 & 0.537 ± 0.021 & 0.832 ± 0.035 & 0.690 ± 0.024 & 0.832 ± 0.021 & 0.828 ± 0.026 & 0.796 ± 0.025 & \textbf{0.876 ± 0.007} & 0.797 ± 0.031 & \underline{0.834 ± 0.032} \\
    \midrule
    \multirow{5}[2]{*}{blood} & 4     & 0.493 ± 0.098 & 0.500 ± 0.116 & 0.500 ± 0.000 & 0.523 ± 0.128 & 0.502 ± 0.138 & 0.562 ± 0.077 & 0.563 ± 0.112 & 0.446 ± 0.069 & 0.561 ± 0.130 & \underline{0.626 ± 0.072} & 0.511 ± 0.011 & 0.530 ± 0.134 & 0.568 ± 0.113 & \textbf{0.691 ± 0.066} \\
          & 8     & 0.542 ± 0.099 & 0.575 ± 0.095 & 0.567 ± 0.060 & 0.564 ± 0.127 & 0.617 ± 0.098 & 0.553 ± 0.084 & 0.604 ± 0.118 & 0.473 ± 0.055 & 0.578 ± 0.109 & \underline{0.648 ± 0.071} & 0.551 ± 0.055 & 0.598 ± 0.121 & 0.618 ± 0.084 & \textbf{0.698 ± 0.050} \\
          & 16    & 0.602 ± 0.069 & 0.635 ± 0.073 & 0.601 ± 0.087 & 0.673 ± 0.089 & 0.593 ± 0.108 & 0.551 ± 0.085 & 0.636 ± 0.080 & 0.497 ± 0.089 & 0.605 ± 0.097 & 0.655 ± 0.063 & 0.523 ± 0.052 & 0.621 ± 0.090 & \underline{0.664 ± 0.071} & \textbf{0.709 ± 0.042} \\
          & 32    & 0.605 ± 0.056 & 0.658 ± 0.065 & 0.682 ± 0.041 & \underline{0.705 ± 0.047} & 0.666 ± 0.047 & 0.571 ± 0.067 & 0.692 ± 0.035 & 0.484 ± 0.078 & 0.644 ± 0.037 & 0.680 ± 0.032 & 0.618 ± 0.063 & 0.695 ± 0.079 & 0.701 ± 0.040 & \textbf{0.723 ± 0.043} \\
          & 48    & 0.626 ± 0.064 & 0.683 ± 0.046 & 0.679 ± 0.034 & 0.686 ± 0.082 & 0.693 ± 0.052 & 0.571 ± 0.086 & \underline{0.713 ± 0.043} & 0.524 ± 0.057 & 0.663 ± 0.058 & 0.699 ± 0.032 & 0.601 ± 0.064 & 0.692 ± 0.056 & 0.702 ± 0.033 & \textbf{0.724 ± 0.049} \\
    \midrule
    \multirow{5}[2]{*}{bank} & 4     & 0.535 ± 0.029 & 0.613 ± 0.050 & 0.500 ± 0.000 & 0.585 ± 0.137 & 0.556 ± 0.070 & 0.531 ± 0.103 & 0.602 ± 0.088 & 0.401 ± 0.156 & 0.609 ± 0.079 & \underline{0.829 ± 0.016} & 0.616 ± 0.053 & 0.723 ± 0.022 & 0.683 ± 0.041 & \textbf{0.837 ± 0.014} \\
          & 8     & 0.608 ± 0.089 & 0.699 ± 0.085 & 0.587 ± 0.095 & 0.676 ± 0.098 & 0.572 ± 0.075 & 0.544 ± 0.075 & 0.706 ± 0.074 & 0.408 ± 0.139 & 0.638 ± 0.058 & \underline{0.830 ± 0.027} & 0.628 ± 0.047 & 0.742 ± 0.021 & 0.723 ± 0.047 & \textbf{0.835 ± 0.013} \\
          & 16    & 0.642 ± 0.071 & 0.730 ± 0.062 & 0.675 ± 0.088 & 0.738 ± 0.060 & 0.584 ± 0.065 & 0.559 ± 0.053 & 0.773 ± 0.084 & 0.447 ± 0.123 & 0.661 ± 0.041 & \underline{0.836 ± 0.020} & 0.616 ± 0.036 & 0.752 ± 0.025 & 0.740 ± 0.040 & \textbf{0.843 ± 0.008} \\
          & 32    & 0.691 ± 0.035 & 0.774 ± 0.033 & 0.752 ± 0.034 & 0.717 ± 0.083 & 0.614 ± 0.048 & 0.608 ± 0.068 & 0.815 ± 0.040 & 0.530 ± 0.080 & 0.674 ± 0.039 & \underline{0.838 ± 0.013} & 0.713 ± 0.029 & 0.768 ± 0.036 & 0.731 ± 0.043 & \textbf{0.841 ± 0.010} \\
          & 48    & 0.709 ± 0.044 & 0.773 ± 0.027 & 0.779 ± 0.030 & 0.771 ± 0.039 & 0.616 ± 0.048 & 0.665 ± 0.037 & 0.829 ± 0.022 & 0.506 ± 0.145 & 0.721 ± 0.074 & \underline{0.838 ± 0.008} & 0.727 ± 0.023 & 0.791 ± 0.031 & 0.732 ± 0.030 & \textbf{0.843 ± 0.011} \\
    \midrule
    \multirow{5}[2]{*}{car} & 4     & 0.562 ± 0.051 & 0.554 ± 0.051 & 0.500 ± 0.000 & 0.580 ± 0.057 & 0.571 ± 0.074 & 0.564 ± 0.026 & 0.612 ± 0.059 & 0.585 ± 0.082 & 0.558 ± 0.043 & 0.840 ± 0.057 & 0.799 ± 0.070 & 0.703 ± 0.060 & 0.656 ± 0.045 & \textbf{0.864 ± 0.034} \\
          & 8     & 0.581 ± 0.061 & 0.643 ± 0.048 & 0.620 ± 0.036 & 0.622 ± 0.033 & 0.622 ± 0.059 & 0.609 ± 0.027 & 0.673 ± 0.060 & 0.660 ± 0.051 & 0.617 ± 0.041 & \textbf{0.918 ± 0.017} & 0.779 ± 0.054 & 0.702 ± 0.072 & 0.704 ± 0.022 & \underline{0.879 ± 0.028} \\
          & 16    & 0.669 ± 0.064 & 0.736 ± 0.036 & 0.679 ± 0.041 & 0.642 ± 0.082 & 0.673 ± 0.061 & 0.652 ± 0.038 & 0.785 ± 0.060 & 0.724 ± 0.025 & 0.628 ± 0.116 & 0.845 ± 0.047 & 0.804 ± 0.023 & 0.744 ± 0.086 & 0.760 ± 0.023 & \textbf{0.881 ± 0.021} \\
          & 32    & 0.745 ± 0.075 & 0.838 ± 0.038 & 0.809 ± 0.038 & 0.707 ± 0.038 & 0.739 ± 0.025 & 0.654 ± 0.036 & \textbf{0.892 ± 0.028} & 0.812 ± 0.040 & 0.782 ± 0.050 & 0.858 ± 0.042 & 0.836 ± 0.022 & 0.837 ± 0.027 & 0.784 ± 0.022 & \underline{0.881 ± 0.029} \\
          & 48    & 0.776 ± 0.051 & 0.853 ± 0.035 & 0.861 ± 0.029 & 0.713 ± 0.032 & 0.778 ± 0.040 & 0.679 ± 0.054 & \textbf{0.935 ± 0.021} & 0.889 ± 0.037 & 0.854 ± 0.038 & 0.834 ± 0.068 & 0.856 ± 0.024 & 0.709 ± 0.131 & 0.805 ± 0.024 & \underline{0.893 ± 0.020} \\
    \midrule
    \multirow{5}[2]{*}{communities} & 4     & 0.568 ± 0.069 & 0.606 ± 0.037 & 0.500 ± 0.000 & 0.559 ± 0.043 & 0.654 ± 0.082 & 0.560 ± 0.030 & 0.687 ± 0.052 & 0.469 ± 0.061 & N/A   & \textbf{0.799 ± 0.013} & 0.686 ± 0.053 & 0.633 ± 0.110 & 0.586 ± 0.071 & \underline{0.784 ± 0.012} \\
          & 8     & 0.603 ± 0.056 & 0.718 ± 0.039 & 0.664 ± 0.060 & 0.586 ± 0.073 & 0.726 ± 0.044 & 0.612 ± 0.053 & 0.743 ± 0.053 & 0.519 ± 0.070 & N/A   & \textbf{0.798 ± 0.025} & 0.725 ± 0.024 & 0.691 ± 0.057 & 0.594 ± 0.095 & \underline{0.782 ± 0.008} \\
          & 16    & 0.623 ± 0.044 & 0.738 ± 0.021 & 0.689 ± 0.072 & 0.639 ± 0.082 & 0.745 ± 0.045 & 0.646 ± 0.033 & 0.769 ± 0.040 & 0.556 ± 0.032 & N/A   & \textbf{0.794 ± 0.028} & 0.725 ± 0.028 & 0.699 ± 0.051 & 0.675 ± 0.071 & \underline{0.789 ± 0.010} \\
          & 32    & 0.655 ± 0.064 & 0.782 ± 0.038 & 0.752 ± 0.038 & 0.727 ± 0.037 & 0.765 ± 0.039 & 0.650 ± 0.030 & \textbf{0.806 ± 0.029} & 0.534 ± 0.049 & N/A   & \underline{0.799 ± 0.023} & 0.668 ± 0.019 & 0.730 ± 0.039 & 0.693 ± 0.105 & 0.781 ± 0.013 \\
          & 48    & 0.689 ± 0.032 & 0.800 ± 0.024 & 0.783 ± 0.027 & 0.742 ± 0.066 & 0.783 ± 0.029 & 0.661 ± 0.028 & \textbf{0.824 ± 0.024} & 0.624 ± 0.103 & N/A   & \underline{0.802 ± 0.016} & 0.637 ± 0.025 & 0.767 ± 0.021 & 0.622 ± 0.094 & 0.779 ± 0.011 \\
    \midrule
    \multirow{5}[2]{*}{credit-g} & 4     & 0.515 ± 0.063 & 0.526 ± 0.056 & 0.500 ± 0.000 & 0.521 ± 0.049 & 0.539 ± 0.072 & 0.508 ± 0.051 & 0.532 ± 0.059 & 0.467 ± 0.077 & 0.580 ± 0.092 & \underline{0.623 ± 0.065} & 0.506 ± 0.018 & 0.498 ± 0.042 & 0.538 ± 0.114 & \textbf{0.669 ± 0.042} \\
          & 8     & 0.543 ± 0.074 & 0.571 ± 0.094 & 0.546 ± 0.064 & 0.600 ± 0.062 & 0.548 ± 0.053 & 0.506 ± 0.059 & 0.584 ± 0.070 & 0.469 ± 0.082 & 0.624 ± 0.110 & \underline{0.627 ± 0.063} & 0.499 ± 0.005 & 0.531 ± 0.053 & 0.588 ± 0.097 & \textbf{0.687 ± 0.053} \\
          & 16    & 0.572 ± 0.068 & 0.599 ± 0.083 & 0.561 ± 0.059 & 0.582 ± 0.075 & 0.558 ± 0.058 & 0.527 ± 0.055 & 0.645 ± 0.085 & 0.477 ± 0.105 & 0.658 ± 0.098 & 0.632 ± 0.072 & 0.513 ± 0.018 & 0.548 ± 0.071 & \underline{0.661 ± 0.059} & \textbf{0.693 ± 0.044} \\
          & 32    & 0.595 ± 0.054 & 0.628 ± 0.062 & 0.635 ± 0.065 & 0.626 ± 0.053 & 0.642 ± 0.058 & 0.515 ± 0.060 & 0.640 ± 0.060 & 0.425 ± 0.142 & \underline{0.685 ± 0.066} & 0.647 ± 0.045 & 0.546 ± 0.038 & 0.569 ± 0.082 & 0.682 ± 0.051 & \textbf{0.709 ± 0.031} \\
          & 48    & 0.620 ± 0.030 & 0.679 ± 0.059 & 0.677 ± 0.046 & 0.669 ± 0.077 & 0.683 ± 0.058 & 0.561 ± 0.028 & 0.693 ± 0.043 & 0.491 ± 0.073 & \underline{0.702 ± 0.059} & 0.646 ± 0.032 & 0.597 ± 0.048 & 0.616 ± 0.090 & 0.670 ± 0.052 & \textbf{0.713 ± 0.045} \\
    \midrule
    \multirow{5}[2]{*}{cdc diabetes} & 4     & 0.539 ± 0.041 & 0.526 ± 0.039 & 0.500 ± 0.000 & 0.565 ± 0.050 & 0.604 ± 0.061 & 0.557 ± 0.035 & 0.605 ± 0.040 & 0.541 ± 0.080 & 0.624 ± 0.013 & \textbf{0.702 ± 0.015} & 0.655 ± 0.008 & 0.604 ± 0.059 & 0.570 ± 0.080 & \underline{0.700 ± 0.012} \\
          & 8     & 0.551 ± 0.041 & 0.605 ± 0.045 & 0.578 ± 0.055 & 0.559 ± 0.051 & 0.622 ± 0.066 & 0.581 ± 0.035 & 0.613 ± 0.051 & 0.578 ± 0.069 & 0.638 ± 0.019 & \textbf{0.703 ± 0.012} & 0.656 ± 0.007 & 0.590 ± 0.073 & 0.609 ± 0.084 & \underline{0.699 ± 0.012} \\
          & 16    & 0.562 ± 0.050 & 0.611 ± 0.056 & 0.600 ± 0.069 & 0.590 ± 0.050 & 0.600 ± 0.069 & 0.572 ± 0.045 & 0.630 ± 0.070 & 0.602 ± 0.048 & 0.620 ± 0.041 & \underline{0.701 ± 0.025} & 0.649 ± 0.016 & 0.633 ± 0.072 & 0.599 ± 0.085 & \textbf{0.705 ± 0.011} \\
          & 32    & 0.596 ± 0.047 & 0.666 ± 0.038 & 0.657 ± 0.034 & 0.624 ± 0.033 & 0.642 ± 0.074 & 0.605 ± 0.025 & 0.691 ± 0.044 & 0.638 ± 0.043 & 0.660 ± 0.027 & \textbf{0.710 ± 0.013} & 0.649 ± 0.011 & 0.673 ± 0.045 & 0.685 ± 0.024 & \underline{0.709 ± 0.005} \\
          & 48    & 0.595 ± 0.031 & 0.672 ± 0.033 & 0.655 ± 0.022 & 0.653 ± 0.026 & 0.667 ± 0.018 & 0.610 ± 0.029 & 0.698 ± 0.030 & 0.630 ± 0.024 & 0.668 ± 0.021 & \textbf{0.709 ± 0.008} & 0.636 ± 0.013 & 0.684 ± 0.029 & 0.675 ± 0.019 & \underline{0.706 ± 0.009} \\
    \midrule
    \multirow{5}[2]{*}{heart} & 4     & 0.595 ± 0.092 & 0.711 ± 0.123 & 0.500 ± 0.000 & 0.510 ± 0.119 & 0.816 ± 0.123 & 0.611 ± 0.062 & 0.767 ± 0.123 & 0.497 ± 0.089 & 0.734 ± 0.109 & 0.858 ± 0.017 & 0.652 ± 0.084 & \underline{0.870 ± 0.024} & 0.602 ± 0.125 & \textbf{0.887 ± 0.022} \\
          & 8     & 0.639 ± 0.104 & 0.721 ± 0.092 & 0.594 ± 0.112 & 0.659 ± 0.081 & 0.867 ± 0.045 & 0.594 ± 0.066 & 0.828 ± 0.051 & 0.521 ± 0.100 & 0.809 ± 0.066 & 0.858 ± 0.025 & 0.563 ± 0.054 & \underline{0.880 ± 0.025} & 0.763 ± 0.073 & \textbf{0.896 ± 0.015} \\
          & 16    & 0.695 ± 0.091 & 0.840 ± 0.040 & 0.804 ± 0.077 & 0.742 ± 0.108 & \underline{0.891 ± 0.034} & 0.632 ± 0.078 & 0.880 ± 0.026 & 0.630 ± 0.090 & 0.848 ± 0.034 & 0.857 ± 0.014 & 0.570 ± 0.056 & 0.877 ± 0.037 & 0.763 ± 0.073 & \textbf{0.898 ± 0.011} \\
          & 32    & 0.748 ± 0.067 & 0.879 ± 0.020 & 0.858 ± 0.017 & 0.794 ± 0.061 & \textbf{0.900 ± 0.028} & 0.643 ± 0.040 & 0.891 ± 0.019 & 0.810 ± 0.029 & 0.862 ± 0.047 & 0.854 ± 0.019 & 0.665 ± 0.048 & \underline{0.892 ± 0.027} & 0.853 ± 0.043 & \underline{0.896 ± 0.013} \\
          & 48    & 0.786 ± 0.056 & 0.884 ± 0.020 & 0.877 ± 0.018 & 0.864 ± 0.032 & \textbf{0.903 ± 0.021} & 0.655 ± 0.046 & 0.898 ± 0.016 & 0.825 ± 0.026 & 0.871 ± 0.036 & 0.853 ± 0.015 & 0.727 ± 0.034 & \underline{0.900 ± 0.017} & 0.854 ± 0.032 & \underline{0.898 ± 0.018} \\
    \midrule
    \multirow{5}[2]{*}{myocardial} & 4     & 0.513 ± 0.038 & 0.519 ± 0.071 & 0.500 ± 0.000 & 0.536 ± 0.061 & 0.551 ± 0.079 & 0.527 ± 0.063 & 0.522 ± 0.074 & 0.516 ± 0.082 & N/A   & \underline{0.609 ± 0.073} & 0.503 ± 0.004 & 0.568 ± 0.070 & 0.580 ± 0.031 & \textbf{0.653 ± 0.048} \\
          & 8     & 0.512 ± 0.047 & 0.537 ± 0.062 & 0.543 ± 0.085 & 0.521 ± 0.078 & 0.513 ± 0.078 & 0.528 ± 0.045 & 0.573 ± 0.102 & 0.538 ± 0.058 & N/A   & \underline{0.616 ± 0.047} & 0.519 ± 0.033 & 0.549 ± 0.067 & 0.583 ± 0.037 & \textbf{0.660 ± 0.046} \\
          & 16    & 0.547 ± 0.060 & 0.592 ± 0.084 & 0.569 ± 0.067 & 0.537 ± 0.098 & 0.519 ± 0.048 & 0.538 ± 0.052 & 0.608 ± 0.059 & 0.587 ± 0.053 & N/A   & \underline{0.632 ± 0.033} & 0.570 ± 0.062 & 0.577 ± 0.062 & 0.576 ± 0.062 & \textbf{0.687 ± 0.048} \\
          & 32    & 0.532 ± 0.072 & 0.559 ± 0.066 & 0.540 ± 0.055 & 0.553 ± 0.057 & 0.545 ± 0.091 & 0.531 ± 0.067 & 0.601 ± 0.078 & 0.591 ± 0.058 & N/A   & \underline{0.625 ± 0.040} & 0.572 ± 0.044 & 0.589 ± 0.074 & 0.587 ± 0.055 & \textbf{0.681 ± 0.055} \\
          & 48    & 0.536 ± 0.061 & 0.561 ± 0.063 & 0.557 ± 0.065 & 0.538 ± 0.077 & 0.559 ± 0.100 & 0.515 ± 0.056 & 0.596 ± 0.086 & 0.583 ± 0.072 & N/A   & \underline{0.625 ± 0.033} & 0.580 ± 0.038 & 0.547 ± 0.066 & 0.539 ± 0.050 & \textbf{0.658 ± 0.034} \\
    \midrule
    \multirow{5}[2]{*}{breast-w} & 4     & 0.836 ± 0.066 & 0.960 ± 0.021 & 0.500 ± 0.000 & 0.783 ± 0.220 & 0.985 ± 0.009 & 0.883 ± 0.137 & 0.986 ± 0.010 & 0.923 ± 0.040 & 0.985 ± 0.010 & 0.974 ± 0.012 & 0.759 ± 0.041 & \underline{0.986 ± 0.008} & 0.979 ± 0.013 & \textbf{0.992 ± 0.006} \\
          & 8     & 0.856 ± 0.080 & 0.974 ± 0.016 & 0.832 ± 0.087 & 0.887 ± 0.114 & 0.983 ± 0.011 & 0.927 ± 0.039 & 0.985 ± 0.009 & 0.926 ± 0.036 & 0.983 ± 0.009 & 0.978 ± 0.012 & 0.768 ± 0.065 & \underline{0.987 ± 0.008} & 0.981 ± 0.015 & \textbf{0.993 ± 0.006} \\
          & 16    & 0.851 ± 0.084 & 0.970 ± 0.020 & 0.876 ± 0.078 & 0.962 ± 0.045 & 0.985 ± 0.010 & 0.921 ± 0.052 & 0.984 ± 0.009 & 0.934 ± 0.029 & 0.979 ± 0.012 & 0.978 ± 0.013 & 0.912 ± 0.051 & \underline{0.988 ± 0.008} & 0.985 ± 0.010 & \textbf{0.993 ± 0.006} \\
          & 32    & 0.910 ± 0.024 & 0.977 ± 0.014 & 0.949 ± 0.035 & 0.971 ± 0.035 & 0.983 ± 0.017 & 0.920 ± 0.043 & \underline{0.988 ± 0.007} & 0.950 ± 0.018 & 0.982 ± 0.013 & 0.977 ± 0.012 & 0.960 ± 0.009 & 0.987 ± 0.007 & 0.988 ± 0.008 & \textbf{0.994 ± 0.004} \\
          & 48    & 0.909 ± 0.048 & 0.966 ± 0.032 & 0.964 ± 0.030 & 0.981 ± 0.014 & 0.987 ± 0.010 & 0.929 ± 0.040 & 0.988 ± 0.006 & 0.949 ± 0.018 & 0.982 ± 0.009 & 0.978 ± 0.012 & 0.976 ± 0.013 & 0.988 ± 0.006 & \underline{0.989 ± 0.006} & \textbf{0.994 ± 0.006} \\
    \midrule
    \multirow{5}[2]{*}{cultivars} & 4     & 0.514 ± 0.073 & 0.493 ± 0.070 & 0.500 ± 0.000 & 0.504 ± 0.064 & 0.494 ± 0.087 & 0.514 ± 0.071 & 0.499 ± 0.091 & 0.445 ± 0.085 & 0.524 ± 0.095 & 0.525 ± 0.047 & 0.531 ± 0.032 & 0.528 ± 0.050 & \underline{0.536 ± 0.076} & \textbf{0.661 ± 0.031} \\
          & 8     & 0.486 ± 0.063 & 0.497 ± 0.062 & 0.503 ± 0.062 & 0.528 ± 0.075 & 0.519 ± 0.095 & 0.498 ± 0.090 & 0.541 ± 0.069 & 0.495 ± 0.056 & 0.554 ± 0.075 & 0.546 ± 0.061 & 0.522 ± 0.066 & \underline{0.561 ± 0.058} & 0.543 ± 0.073 & \textbf{0.653 ± 0.025} \\
          & 16    & 0.495 ± 0.066 & 0.523 ± 0.076 & 0.510 ± 0.042 & 0.530 ± 0.055 & 0.527 ± 0.082 & 0.519 ± 0.093 & 0.511 ± 0.071 & 0.491 ± 0.110 & \underline{0.584 ± 0.084} & 0.531 ± 0.092 & 0.563 ± 0.070 & 0.582 ± 0.063 & 0.576 ± 0.043 & \textbf{0.679 ± 0.027} \\
          & 32    & 0.536 ± 0.083 & 0.525 ± 0.055 & 0.531 ± 0.049 & 0.527 ± 0.050 & 0.525 ± 0.071 & 0.540 ± 0.063 & 0.530 ± 0.083 & 0.517 ± 0.099 & 0.591 ± 0.062 & 0.527 ± 0.088 & 0.564 ± 0.068 & 0.568 ± 0.075 & \underline{0.622 ± 0.057} & \textbf{0.676 ± 0.041} \\
          & 48    & 0.547 ± 0.094 & 0.570 ± 0.067 & 0.546 ± 0.065 & 0.518 ± 0.052 & 0.519 ± 0.076 & 0.555 ± 0.070 & 0.555 ± 0.094 & 0.558 ± 0.095 & 0.608 ± 0.039 & 0.557 ± 0.082 & 0.568 ± 0.038 & 0.579 ± 0.094 & \underline{0.615 ± 0.049} & \textbf{0.670 ± 0.049} \\
    \midrule
    \multirow{5}[2]{*}{NHANES} & 4     & 0.662 ± 0.183 & 0.805 ± 0.152 & 0.500 ± 0.000 & 0.870 ± 0.200 & 0.773 ± 0.162 & 0.512 ± 0.018 & 0.898 ± 0.098 & 0.457 ± 0.124 & \underline{0.975 ± 0.062} & 0.525 ± 0.047 & 0.531 ± 0.032 & 0.527 ± 0.189 & 0.673 ± 0.212 & \textbf{0.987 ± 0.010} \\
          & 8     & 0.962 ± 0.031 & 0.900 ± 0.137 & 0.880 ± 0.135 & 0.968 ± 0.034 & 0.812 ± 0.119 & 0.518 ± 0.016 & 0.976 ± 0.043 & 0.519 ± 0.114 & \textbf{0.999 ± 0.001} & 0.546 ± 0.061 & 0.522 ± 0.066 & 0.750 ± 0.116 & 0.925 ± 0.037 & \underline{0.998 ± 0.002} \\
          & 16    & 0.967 ± 0.023 & 0.974 ± 0.020 & 0.956 ± 0.039 & 0.988 ± 0.011 & 0.844 ± 0.116 & 0.532 ± 0.061 & \underline{0.999 ± 0.001} & 0.521 ± 0.058 & 0.999 ± 0.002 & 0.531 ± 0.092 & 0.563 ± 0.070 & 0.899 ± 0.082 & 0.969 ± 0.018 & \textbf{0.999 ± 0.001} \\
          & 32    & 0.988 ± 0.016 & 0.990 ± 0.012 & 0.989 ± 0.028 & 0.987 ± 0.011 & 0.835 ± 0.142 & 0.535 ± 0.069 & \textbf{1.000 ± 0.000} & 0.792 ± 0.073 & \underline{1.000 ± 0.000} & 0.527 ± 0.088 & 0.564 ± 0.068 & 0.960 ± 0.023 & 0.959 ± 0.012 & 0.999 ± 0.001 \\
          & 48    & 0.989 ± 0.015 & 0.992 ± 0.013 & 0.998 ± 0.007 & 0.990 ± 0.009 & 0.810 ± 0.084 & 0.579 ± 0.103 & \textbf{1.000 ± 0.000} & 0.840 ± 0.092 & 0.999 ± 0.002 & 0.557 ± 0.082 & 0.568 ± 0.038 & 0.977 ± 0.026 & 0.967 ± 0.008 & \underline{0.999 ± 0.001} \\
    \midrule
    \multirow{5}[2]{*}{gallstone} & 4     & 0.559 ± 0.090 & 0.564 ± 0.085 & 0.500 ± 0.000 & 0.493 ± 0.064 & 0.505 ± 0.103 & 0.509 ± 0.053 & \underline{0.570 ± 0.095} & 0.440 ± 0.081 & 0.465 ± 0.071 & 0.565 ± 0.071 & 0.533 ± 0.046 & 0.540 ± 0.078 & 0.471 ± 0.076 & \textbf{0.658 ± 0.042} \\
          & 8     & 0.555 ± 0.094 & 0.571 ± 0.115 & 0.508 ± 0.089 & 0.532 ± 0.088 & 0.522 ± 0.097 & 0.505 ± 0.058 & \underline{0.597 ± 0.100} & 0.432 ± 0.070 & 0.473 ± 0.102 & 0.558 ± 0.074 & 0.543 ± 0.043 & 0.532 ± 0.122 & 0.436 ± 0.067 & \textbf{0.653 ± 0.054} \\
          & 16    & 0.592 ± 0.099 & 0.591 ± 0.101 & 0.589 ± 0.133 & 0.596 ± 0.102 & 0.537 ± 0.128 & 0.528 ± 0.091 & \underline{0.612 ± 0.122} & 0.426 ± 0.071 & 0.461 ± 0.077 & 0.563 ± 0.112 & 0.529 ± 0.062 & 0.557 ± 0.083 & 0.481 ± 0.040 & \textbf{0.662 ± 0.045} \\
          & 32    & 0.598 ± 0.071 & \underline{0.668 ± 0.084} & 0.665 ± 0.086 & 0.667 ± 0.093 & 0.559 ± 0.074 & 0.558 ± 0.076 & \textbf{0.696 ± 0.103} & 0.454 ± 0.077 & 0.559 ± 0.053 & 0.631 ± 0.079 & 0.551 ± 0.040 & 0.613 ± 0.077 & 0.485 ± 0.074 & 0.659 ± 0.045 \\
          & 48    & 0.663 ± 0.054 & 0.725 ± 0.076 & \underline{0.740 ± 0.050} & 0.711 ± 0.087 & 0.598 ± 0.080 & 0.556 ± 0.092 & \textbf{0.793 ± 0.052} & 0.436 ± 0.098 & 0.548 ± 0.077 & 0.628 ± 0.087 & 0.546 ± 0.061 & 0.661 ± 0.053 & 0.499 ± 0.067 & 0.663 ± 0.054 \\
    \bottomrule
    \end{tabular}%
    }
  \label{tab:cls_appendix}%
\end{table*}%

\begin{table*}[htbp]
    \centering
    \caption{Performance comparison of traditional methods and LLM-based methods across various datasets and 4-, 8-, 16-, 32-, and 48-shot settings for regression tasks. The best-performing method is bolded, and the second-best is underlined. Each value represents the NRMSE score, reported as the mean ± standard deviation across 10 random seeds.}
      \resizebox{\linewidth}{!}{%
    \begin{tabular}{l|c|cccccccccc}
    \toprule
    \multirow{3}[6]{*}{Dataset} & \multirow{3}[6]{*}{Shot} & \multicolumn{6}{c|}{Traditional Method}       & \multicolumn{4}{c}{LLM-based Method \newline{}} \\
\cmidrule{3-12}          &       & \multicolumn{6}{c|}{Training Required}        & \multicolumn{1}{c|}{\makecell[c]{Fine-tuned \\Required}} & \multicolumn{3}{c}{No Fine-tuning Required} \\
\cmidrule{3-12}          &       & ElasticNet & MLP   & CART  & \multicolumn{1}{c}{\makecell[c]{Random\\Forest}} & XGBoost & TabPFN & TP-BERTa & LIFT  & P2T   & Ours \\
    \midrule
    \multirow{5}[2]{*}{bike} & 4     & 0.266 ± 0.080 & 0.225 ± 0.038 & 0.220 ± 0.026 & 0.194 ± 0.025 & 0.224 ± 0.055 & 0.192 ± 0.013 & 0.264 ± 0.004 & \underline{0.161 ± 0.011} & 0.190 ± 0.020 & \textbf{0.156 ± 0.007} \\
          & 8     & 0.211 ± 0.065 & 0.247 ± 0.082 & 0.300 ± 0.086 & 0.197 ± 0.024 & 0.243 ± 0.121 & 0.191 ± 0.028 & 0.262 ± 0.004 & 0.259 ± 0.058 & \underline{0.185 ± 0.013} & \textbf{0.149 ± 0.007} \\
          & 16    & 0.202 ± 0.053 & 0.256 ± 0.066 & 0.304 ± 0.132 & 0.200 ± 0.068 & 0.193 ± 0.024 & 0.165 ± 0.012 & 0.261 ± 0.004 & \underline{0.156 ± 0.015} & 0.194 ± 0.014 & \textbf{0.144 ± 0.011} \\
          & 32    & 0.205 ± 0.033 & 0.264 ± 0.106 & 0.200 ± 0.085 & 0.170 ± 0.077 & 0.184 ± 0.084 & \textbf{0.140 ± 0.053} & 0.259 ± 0.004 & 0.152 ± 0.015 & 0.165 ± 0.060 & \underline{0.142 ± 0.006} \\
          & 48    & 0.184 ± 0.016 & 0.208 ± 0.076 & 0.215 ± 0.101 & 0.153 ± 0.059 & 0.153 ± 0.055 & \textbf{0.126 ± 0.047} & 0.258 ± 0.004 & 0.155 ± 0.020 & 0.176 ± 0.009 & \underline{0.141 ± 0.007} \\
    \midrule
    \multirow{5}[2]{*}{cpu\_small} & 4     & 0.322 ± 0.157 & 49.042 ± 82.905 & 0.168 ± 0.050 & 0.180 ± 0.017 & 0.198 ± 0.019 & \underline{0.167 ± 0.024} & 0.789 ± 0.003 & 0.173 ± 0.082 & 0.234 ± 0.071 & \textbf{0.161 ± 0.038} \\
          & 8     & 0.389 ± 0.314 & 9.865 ± 27.918 & 0.172 ± 0.019 & 0.167 ± 0.022 & 0.187 ± 0.009 & \underline{0.164 ± 0.029} & 0.770 ± 0.004 & 0.172 ± 0.036 & 0.167 ± 0.022 & \textbf{0.154 ± 0.024} \\
          & 16    & 0.335 ± 0.437 & 0.987 ± 0.375 & 0.156 ± 0.067 & 0.122 ± 0.040 & 0.180 ± 0.037 & 0.118 ± 0.039 & 0.745 ± 0.004 & \textbf{0.101 ± 0.045} & 0.147 ± 0.036 & \underline{0.114 ± 0.027} \\
          & 32    & 0.234 ± 0.146 & 1.674 ± 2.491 & 0.101 ± 0.047 & \underline{0.098 ± 0.048} & 0.151 ± 0.057 & 0.114 ± 0.060 & 0.720 ± 0.004 & \textbf{0.098 ± 0.048} & 0.101 ± 0.043 & 0.116 ± 0.022 \\
          & 48    & 0.230 ± 0.142 & 0.867 ± 0.003 & \textbf{0.084 ± 0.051} & 0.094 ± 0.047 & 0.129 ± 0.053 & 0.129 ± 0.056 & 0.704 ± 0.004 & 0.090 ± 0.052 & \underline{0.090 ± 0.040} & 0.122 ± 0.021 \\
    \midrule
    \multirow{5}[2]{*}{diamonds} & 4     & 0.142 ± 0.063 & 0.153 ± 0.038 & 0.152 ± 0.020 & 0.152 ± 0.021 & 0.156 ± 0.019 & 0.157 ± 0.078 & 0.302 ± 0.002 & \textbf{0.080 ± 0.009} & 0.116 ± 0.021 & \underline{0.090 ± 0.010} \\
          & 8     & 0.169 ± 0.053 & 0.116 ± 0.026 & 0.150 ± 0.057 & 0.118 ± 0.019 & 0.129 ± 0.014 & 0.153 ± 0.103 & 0.302 ± 0.002 & \textbf{0.080 ± 0.007} & 0.109 ± 0.014 & \underline{0.095 ± 0.013} \\
          & 16    & 0.123 ± 0.031 & 0.115 ± 0.028 & 0.126 ± 0.019 & 0.120 ± 0.015 & 0.134 ± 0.021 & 0.119 ± 0.059 & 0.302 ± 0.002 & \textbf{0.082 ± 0.011} & 0.109 ± 0.008 & \underline{0.095 ± 0.007} \\
          & 32    & 0.092 ± 0.013 & 0.101 ± 0.029 & 0.116 ± 0.017 & 0.096 ± 0.009 & 0.107 ± 0.009 & \underline{0.089 ± 0.020} & 0.302 ± 0.002 & \textbf{0.084 ± 0.013} & 0.095 ± 0.006 & 0.090 ± 0.005 \\
          & 48    & \underline{0.087 ± 0.009} & 0.094 ± 0.037 & 0.112 ± 0.017 & 0.095 ± 0.010 & 0.097 ± 0.010 & 0.089 ± 0.035 & 0.302 ± 0.002 & \textbf{0.078 ± 0.011} & 0.092 ± 0.006 & 0.090 ± 0.005 \\
    \midrule
    \multirow{5}[2]{*}{forest-fires} & 4     & 0.128 ± 0.031 & 0.130 ± 0.031 & 0.127 ± 0.029 & 0.127 ± 0.030 & 0.127 ± 0.030 & \underline{0.127 ± 0.029} & 0.128 ± 0.030 & 0.127 ± 0.030 & 0.127 ± 0.030 & \textbf{0.126 ± 0.029} \\
          & 8     & 0.170 ± 0.109 & 0.128 ± 0.030 & 0.204 ± 0.221 & 0.143 ± 0.042 & 0.140 ± 0.036 & 0.170 ± 0.131 & \underline{0.127 ± 0.029} & 0.153 ± 0.071 & 0.136 ± 0.033 & \textbf{0.126 ± 0.029} \\
          & 16    & 0.147 ± 0.051 & 0.133 ± 0.034 & 0.214 ± 0.192 & 0.146 ± 0.046 & 0.144 ± 0.044 & 0.136 ± 0.037 & \underline{0.126 ± 0.029} & 0.158 ± 0.070 & 0.143 ± 0.045 & \textbf{0.123 ± 0.029} \\
          & 32    & 0.127 ± 0.027 & 0.132 ± 0.029 & 0.162 ± 0.079 & 0.139 ± 0.037 & 0.128 ± 0.030 & 0.125 ± 0.029 & 0.126 ± 0.030 & 0.149 ± 0.053 & \textbf{0.121 ± 0.054} & \underline{0.122 ± 0.029} \\
          & 48    & 0.126 ± 0.027 & 0.152 ± 0.072 & 0.197 ± 0.112 & 0.136 ± 0.037 & 0.136 ± 0.031 & 0.125 ± 0.028 & 0.126 ± 0.029 & 0.160 ± 0.052 & \underline{0.119 ± 0.052} & \textbf{0.118 ± 0.031} \\
    \midrule
    \multirow{5}[2]{*}{houses} & 4     & 0.393 ± 0.232 & 6.056 ± 2.924 & 0.252 ± 0.044 & 0.180 ± 0.034 & \underline{0.156 ± 0.020} & 0.176 ± 0.023 & 1.270 ± 0.143 & 0.214 ± 0.085 & 0.358 ± 0.209 & \textbf{0.150 ± 0.030} \\
          & 8     & 0.286 ± 0.211 & 6.218 ± 6.041 & 0.181 ± 0.028 & 0.170 ± 0.015 & 0.161 ± 0.023 & \underline{0.142 ± 0.023} & 0.801 ± 0.108 & 0.169 ± 0.032 & 0.192 ± 0.052 & \textbf{0.131 ± 0.011} \\
          & 16    & 0.183 ± 0.037 & 2.840 ± 1.398 & 0.175 ± 0.025 & 0.150 ± 0.012 & 0.156 ± 0.015 & \underline{0.131 ± 0.012} & 0.285 ± 0.059 & 0.148 ± 0.013 & 0.163 ± 0.018 & \textbf{0.128 ± 0.011} \\
          & 32    & 0.134 ± 0.024 & 3.275 ± 4.016 & 0.174 ± 0.025 & 0.135 ± 0.017 & 0.142 ± 0.020 & \textbf{0.112 ± 0.013} & 0.172 ± 0.014 & 0.153 ± 0.021 & 0.149 ± 0.014 & \underline{0.127 ± 0.011} \\
          & 48    & \underline{0.121 ± 0.018} & 2.081 ± 2.099 & 0.162 ± 0.022 & 0.131 ± 0.012 & 0.131 ± 0.016 & \textbf{0.104 ± 0.012} & 0.172 ± 0.014 & 0.156 ± 0.025 & 0.141 ± 0.011 & 0.126 ± 0.011 \\
    \midrule
    \multirow{5}[2]{*}{insurance} & 4     & 0.240 ± 0.044 & 0.235 ± 0.056 & 0.282 ± 0.070 & 0.236 ± 0.035 & 0.236 ± 0.027 & 0.194 ± 0.041 & 0.325 ± 0.039 & \underline{0.125 ± 0.018} & 0.186 ± 0.055 & \textbf{0.115 ± 0.021} \\
          & 8     & 0.224 ± 0.045 & 0.207 ± 0.023 & 0.295 ± 0.126 & 0.255 ± 0.065 & 0.245 ± 0.042 & 0.163 ± 0.038 & 0.325 ± 0.039 & \underline{0.116 ± 0.020} & 0.145 ± 0.026 & \textbf{0.107 ± 0.014} \\
          & 16    & 0.185 ± 0.017 & 0.193 ± 0.012 & 0.230 ± 0.049 & 0.232 ± 0.028 & 0.234 ± 0.031 & 0.158 ± 0.048 & 0.325 ± 0.039 & \underline{0.116 ± 0.017} & 0.119 ± 0.044 & \textbf{0.105 ± 0.012} \\
          & 32    & 0.156 ± 0.023 & 0.158 ± 0.021 & 0.181 ± 0.086 & 0.212 ± 0.079 & 0.207 ± 0.077 & 0.117 ± 0.045 & 0.325 ± 0.039 & \underline{0.112 ± 0.015} & 0.119 ± 0.013 & \textbf{0.100 ± 0.010} \\
          & 48    & 0.153 ± 0.020 & 0.162 ± 0.030 & 0.222 ± 0.056 & 0.236 ± 0.030 & 0.233 ± 0.037 & 0.122 ± 0.017 & 0.325 ± 0.039 & \underline{0.112 ± 0.016} & 0.114 ± 0.042 & \textbf{0.102 ± 0.014} \\
    \midrule
    \multirow{5}[2]{*}{plasma\_retinol} & 4     & 0.490 ± 0.329 & 0.396 ± 0.112 & 0.292 ± 0.100 & 0.239 ± 0.049 & \textbf{0.215 ± 0.028} & 0.227 ± 0.046 & 0.633 ± 0.140 & 0.331 ± 0.112 & 0.412 ± 0.163 & \underline{0.216 ± 0.038} \\
          & 8     & 0.388 ± 0.101 & 0.472 ± 0.248 & 0.304 ± 0.097 & 0.240 ± 0.063 & \underline{0.234 ± 0.048} & 0.242 ± 0.060 & 0.631 ± 0.140 & 0.340 ± 0.108 & 0.316 ± 0.076 & \textbf{0.203 ± 0.028} \\
          & 16    & 0.301 ± 0.171 & 0.445 ± 0.371 & 0.296 ± 0.080 & 0.217 ± 0.038 & 0.220 ± 0.025 & \underline{0.203 ± 0.023} & 0.628 ± 0.139 & 0.340 ± 0.129 & 0.268 ± 0.064 & \textbf{0.192 ± 0.018} \\
          & 32    & 0.315 ± 0.281 & 0.439 ± 0.410 & 0.307 ± 0.084 & 0.231 ± 0.040 & 0.215 ± 0.032 & \underline{0.203 ± 0.024} & 0.626 ± 0.138 & 0.327 ± 0.079 & 0.270 ± 0.053 & \textbf{0.193 ± 0.017} \\
          & 48    & 0.216 ± 0.031 & 0.389 ± 0.268 & 0.287 ± 0.083 & 0.214 ± 0.025 & 0.210 ± 0.029 & \underline{0.202 ± 0.025} & 0.624 ± 0.138 & 0.309 ± 0.051 & 0.268 ± 0.053 & \textbf{0.193 ± 0.019} \\
    \midrule
    \multirow{5}[2]{*}{wine} & 4     & 0.206 ± 0.065 & 0.361 ± 0.185 & 0.209 ± 0.030 & 0.170 ± 0.027 & 0.169 ± 0.030 & 0.166 ± 0.029 & \underline{0.150 ± 0.005} & 0.164 ± 0.023 & 0.252 ± 0.068 & \textbf{0.141 ± 0.010} \\
          & 8     & 0.193 ± 0.042 & 0.228 ± 0.077 & 0.188 ± 0.033 & 0.167 ± 0.032 & 0.161 ± 0.024 & 0.153 ± 0.023 & \underline{0.146 ± 0.003} & 0.158 ± 0.010 & 0.183 ± 0.026 & \textbf{0.137 ± 0.006} \\
          & 16    & 0.178 ± 0.032 & 0.207 ± 0.060 & 0.187 ± 0.021 & 0.156 ± 0.018 & 0.158 ± 0.016 & 0.155 ± 0.023 & \underline{0.148 ± 0.004} & 0.158 ± 0.011 & 0.161 ± 0.018 & \textbf{0.132 ± 0.002} \\
          & 32    & 0.157 ± 0.022 & 0.170 ± 0.024 & 0.186 ± 0.009 & 0.146 ± 0.007 & 0.149 ± 0.009 & 0.137 ± 0.006 & 0.154 ± 0.012 & 0.158 ± 0.011 & \underline{0.135 ± 0.047} & \textbf{0.131 ± 0.002} \\
          & 48    & 0.143 ± 0.010 & 0.161 ± 0.016 & 0.178 ± 0.017 & 0.141 ± 0.006 & 0.141 ± 0.006 & 0.135 ± 0.006 & 0.151 ± 0.009 & 0.159 ± 0.007 & \underline{0.133 ± 0.047} & \textbf{0.131 ± 0.003} \\
    \midrule
    \multirow{5}[2]{*}{cultivars} & 4     & 0.325 ± 0.143 & 0.314 ± 0.056 & 0.268 ± 0.056 & 0.234 ± 0.024 & \underline{0.218 ± 0.023} & 0.300 ± 0.240 & 1.403 ± 0.151 & 0.370 ± 0.131 & 0.328 ± 0.103 & \textbf{0.206 ± 0.021} \\
          & 8     & 0.369 ± 0.221 & 0.330 ± 0.097 & 0.263 ± 0.039 & 0.235 ± 0.049 & 0.224 ± 0.039 & \underline{0.221 ± 0.040} & 1.402 ± 0.151 & 0.321 ± 0.121 & 0.288 ± 0.085 & \textbf{0.196 ± 0.017} \\
          & 16    & 0.224 ± 0.034 & 0.369 ± 0.284 & 0.277 ± 0.042 & 0.234 ± 0.035 & 0.231 ± 0.041 & \underline{0.217 ± 0.032} & 1.400 ± 0.151 & 0.270 ± 0.074 & 0.266 ± 0.077 & \textbf{0.198 ± 0.023} \\
          & 32    & 0.214 ± 0.025 & 0.248 ± 0.043 & 0.273 ± 0.041 & 0.216 ± 0.021 & 0.223 ± 0.025 & \underline{0.211 ± 0.024} & 1.398 ± 0.151 & 0.266 ± 0.065 & 0.259 ± 0.086 & \textbf{0.193 ± 0.019} \\
          & 48    & 0.209 ± 0.023 & 0.227 ± 0.029 & 0.260 ± 0.039 & 0.212 ± 0.025 & 0.212 ± 0.021 & \underline{0.206 ± 0.023} & 1.396 ± 0.151 & 0.267 ± 0.062 & 0.272 ± 0.099 & \textbf{0.191 ± 0.022} \\
    \midrule
    \multirow{5}[2]{*}{\makecell[l]{infrared\_thermography\_\\temperature}} & 4     & 0.152 ± 0.025 & 0.288 ± 0.148 & 0.130 ± 0.025 & 0.116 ± 0.018 & 0.151 ± 0.021 & 0.138 ± 0.057 & 6.628 ± 0.707 & \underline{0.129 ± 0.017} & 1.637 ± 1.137 & \textbf{0.108 ± 0.012} \\
          & 8     & 0.132 ± 0.042 & 0.138 ± 0.049 & 0.129 ± 0.016 & 0.109 ± 0.022 & 0.160 ± 0.049 & 0.110 ± 0.024 & 5.898 ± 0.628 & \underline{0.105 ± 0.013} & 0.722 ± 1.062 & \textbf{0.102 ± 0.010} \\
          & 16    & 0.120 ± 0.048 & 0.140 ± 0.049 & 0.117 ± 0.017 & \underline{0.099 ± 0.012} & 0.109 ± 0.016 & 0.112 ± 0.058 & 5.036 ± 0.536 & \textbf{0.098 ± 0.011} & 0.106 ± 0.014 & 0.103 ± 0.008 \\
          & 32    & 0.109 ± 0.085 & 0.121 ± 0.037 & 0.103 ± 0.012 & \textbf{0.085 ± 0.012} & \underline{0.093 ± 0.009} & 0.093 ± 0.046 & 3.516 ± 0.375 & 0.095 ± 0.014 & 0.096 ± 0.011 & 0.097 ± 0.007 \\
          & 48    & \underline{0.079 ± 0.006} & 0.437 ± 0.619 & 0.103 ± 0.015 & 0.084 ± 0.010 & 0.088 ± 0.009 & \textbf{0.078 ± 0.015} & 1.842 ± 0.678 & 0.096 ± 0.012 & 0.140 ± 0.152 & 0.096 ± 0.009 \\
    \bottomrule
    \end{tabular}%
    }
  \label{tab:reg_appendix}%
\end{table*}%

\end{document}